\documentclass[10pt,twocolumn,letterpaper]{article}
\usepackage[pagenumbers]{cvpr} 

\usepackage[table,usenames,dvipsnames]{xcolor}
\usepackage{amsmath, amssymb, amsthm}
\usepackage{graphicx}
\usepackage{enumitem}
\usepackage{appendix}
\usepackage{booktabs}
\usepackage{microtype}  
\usepackage{nicefrac}
\usepackage{wrapfig}
\usepackage{pifont}
\graphicspath{{figs/}}
\usepackage{adjustbox}

\usepackage{amsmath,amsfonts,bm}

\def\eqref#1{equation~\ref{#1}}

\def\1{\bm{1}}

\def\vs{{\bm{s}}}

\DeclareMathAlphabet{\mathsfit}{\encodingdefault}{\sfdefault}{m}{sl}
\SetMathAlphabet{\mathsfit}{bold}{\encodingdefault}{\sfdefault}{bx}{n}

\usepackage{tcolorbox}

\newcommand{\seqmodname}{Kinaema}

\newcommand{\ours}{Chimera}

\newcommand{\obs}{\textbf{o}}
\newcommand{\query}{q}

\newcommand{\mem}{\mathbf{M}}
\newcommand{\bott}{\mathbf{B}}
\newcommand{\bottt}{\tilde{\mathbf{B}}}

\newcommand{\trmapping}{\overset{\theta}{\rightarrow}}
\newcommand{\trmappingl}{\overset{\theta}{\longrightarrow}}

\definecolor{ColComments}{HTML}{F7C5A6}

\newtcolorbox{highlightbox}{
    colback=yellow!10, 
    colframe=white,    
    sharp corners, 
    boxrule=0pt, 
    enhanced, 
    breakable,
    left=2pt, right=2pt, top=2pt, bottom=2pt 
}

\newtcolorbox{highlightitem}{
    enhanced,
    breakable,
    colback=yellow!10,
    colframe=white,
    sharp corners,
    boxrule=0pt,
    before upper={\list{}{\leftmargin=0em\itemindent=0em}\item},
    after upper={\endlist},
    left=20pt, 
    right=5pt,
    top=0pt,
    bottom=2pt,
    before skip=10pt,
    after skip=10pt
}

\DeclareMathOperator{\Encode}{\it Enc}

\DeclareMathOperator{\Encodegoal}{\Encode_{\text{\em goal}}}

\DeclareMathOperator{\Update}{\it Update}

\DeclareMathOperator{\Linear}{\it Linear}

\DeclareMathOperator{\SA}{\it SelfAttn}

\DeclareMathOperator{\GRU}{\it GRU}

\newcommand{\tcls}{\texttt{CLS}\xspace}

\newcommand{\hmem}{\mathbf{M}}

\newcommand{\yes}{\textcolor{black}{\ding{51}}}
\newcommand{\no}{\textcolor{black}{\ding{55}}}

\makeatletter
\DeclareRobustCommand\onedot{\futurelet\@let@token\@onedot}
\def\@onedot{\ifx\@let@token.\else.\null\fi\xspace}
\def\eg{\emph{e.g}\onedot} 
\def\ie{\emph{i.e}\onedot} 
\def\cf{\emph{cf}\onedot} 
 \def\vs{\emph{vs}\onedot}
\def\wrt{w.r.t\onedot}

\makeatother

\newcommand\mydots{\makebox[0.6em][c]{.\hfil.\hfil.}}
\newcommand{\myparagraph}[1]{\noindent \textbf{#1}}
\newcommand{\metrics}{%
{$\genfrac{}{}{0pt}{}{1m}{10^o}$}&
{$\genfrac{}{}{0pt}{}{1m}{90^o}$}&
{$\genfrac{}{}{0pt}{}{2m}{90^o}$}
}

\newcolumntype{C}[1]{>{\columncolor{#1}}c}

\definecolor{cvprblue}{rgb}{0.21,0.49,0.74}
\definecolor{TableGray1}{HTML}{D4D4D4}
\definecolor{TableGray2}{HTML}{E6E6E6}
\definecolor{TableGray3}{HTML}{EEEEEE}

\definecolor{T200}{HTML}{D09CB5}
\definecolor{T800}{HTML}{E7CBD7}
\definecolor{T200T}{HTML}{CBF0C0}
\definecolor{T800T}{HTML}{E2F0DD}

\definecolor{colsee}{HTML}{D7E6F5}
\definecolor{colun}{HTML}{B4D5F4}
\definecolor{colov}{HTML}{B4C0F4}

\definecolor{ColComments}{HTML}{F7C5A6}

\newcommand{\rpevalbox}[1]{\tcbox[on line,colframe=white,boxsep=0pt,left=1pt,right=1pt,top=0pt,bottom=0pt,colback=T800]{#1}}
\newcommand{\rpetestbox}[1]{\tcbox[on line,colframe=white,boxsep=0pt,left=1pt,right=1pt,top=0pt,bottom=0pt,colback=T800T]{#1}}

\newcommand{\trainbox}[1]{\tcbox[on line,colframe=white,boxsep=0pt,left=1pt,right=1pt,top=0pt,bottom=0pt,colback=TableGray2]{#1}}

\newcommand{\rpetrain}{\trainbox{\textbf{RPE-train}}}
\newcommand{\rpeval}{\rpevalbox{\textbf{RPE-val}}}
\newcommand{\rpetest}{\rpetestbox{\textbf{RPE-test}}}

\usepackage{tikz}
\usetikzlibrary{arrows.meta, positioning, chains, shapes, shadows.blur}
\tikzset{
 flow_box/.style={rectangle, draw, thick, fill=black!5, minimum height=1.5cm, text centered, font=\bfseries, align=center},
 io/.style={rectangle, rounded corners, draw, thick, fill=blue!10, minimum size=0.6cm, align=center},
 output/.style={rectangle, rounded corners, draw, thick, fill=green!10, minimum size=0.6cm, align=center},
 arrow/.style={-Latex, thick, shorten >=1pt, shorten <=1pt}
}

\definecolor{cvprblue}{rgb}{0.21,0.49,0.74}
\usepackage[pagebackref,breaklinks,colorlinks,allcolors=cvprblue]{hyperref}

\def\confName{CVPR}
\def\confYear{2027}

\title{Supervising Recurrent Vision Transformers from Bottlenecked Context}

\author{Philippe Weinzaepfel\thanks{Equal contribution} ~~~~ Christian Wolf\footnotemark[1] ~~~~  Mert Bülent Sariyildiz \\
  Guillaume Bono ~~~~ Gianluca Monaci \\[1mm]
  NAVER LABS Europe \\[1mm]
  \url{https://europe.naverlabs.com/chimera}
}

\begin{document}

\maketitle

\begin{abstract}
Transformers are AI's workhorse 
but their computational cost becomes prohibitive when processing long sequences. We target long-horizon streaming vision and robotics applications, where it is particularly impractical to store and maintain a history of observations. Recurrent Transformers  address this limitation by maintaining fixed-size memory but their performance lags behind that of transformers operating over the full observation history. We argue that this gap does not stem from architectural limitations, but from differences in how these models learn to compress past information. Without access to an observation history, recurrent models must explicitly decide what to retain in memory at each step, a significantly harder learning problem. In this work, we propose a distillation approach that transfers the compression strategy of a classical full-history transformer to a recurrent variant. We enable this by designing a teacher model that explicitly compresses its observation history into a fixed-size bottleneck representation and  directly supervise the student's memory with this bottleneck representation, effectively aligning the two compression mechanisms. We show that this approach enables training of a recurrent latent robotic memory with linear-time complexity on the Mem-RPE task while substantially narrowing the performance gap to full-history transformers. We additionally validate the same principle on streaming visual question answering (VQA) and observe improved recurrent predictions thanks to memory distillation.
\end{abstract}

\begin{figure}
    \centering
    \includegraphics[width=\linewidth]{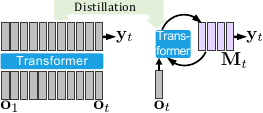}
    \caption{\label{fig:teaser}Transformers (left) attend over the full history of observations with complexity $O(t^2)$ \underline{at each step $t$}. Recurrent transformers (right) maintain memory and take decisions with $O(1)$. We distill ``\textit{History Transformers}'' into recurrent transformers, combining prediction performance and efficiency.}
\end{figure}


\section{Introduction}
\label{sec:intro}

Transformers \citep{vaswani2017attention} successfully emerged as arguably the most general model architecture in Machine Learning (ML) history and have been successfully applied in NLP \citep{vaswani2017attention,gpt32020}, computer vision \citep{dosovitskiy2021an,wang2024dust3r}, speech \citep{whisper2023}, robotics \citep{pi05}, physics \citep{janny2023eagle,xu2025amrtransformer} 
and more. Their quadratic complexity $O(T^2$) \wrt. the number of input embeddings $T$ 
is well known to be a shortcoming and this 
has posed some stringent problems for domains like robotics, where attending to a rapidly growing number of observations can become impossible, \cf~\Cref{fig:teaser} (left).
While it is possible to ignore this in applications like manipulation, where often past observations beyond a  recent cut-off point are ignored (treating them as fully observable problems), this is not an option for applications where the full past needs to be remembered. Examples are 3D reconstruction \citep{wang2024dust3r,vggt2025}, SLAM \citep{vggtslam2_0}, navigation towards previously observed goals \citep{kinaema2025} and generally any application where agents need to be able to recall any previously observed information. 
Our main target application is memory-based pose estimation, where an agent is asked to situate a visual query in the explored environment and with respect to its own moving coordinate frame: ``\textit{Where is this thing with respect to your current position?}'' Answering this question requires information beyond the current observation.

Recurrent models like LSTMs \citep{DBLP:journals/neco/HochreiterS97} and GRUs \citep{cho-etal-2014-learning} have potentially unlimited context length and are restricted only by the size of their memory, the hidden state.
However, while they initially were the models of choice in NLP, they famously lacked scaling capabilities and were largely replaced by transformers.
Recurrent models have recently made a comeback by replacing the flat vectorial hidden state with a memory bank and performing the update with a transformer \citep{moog2024,kinaema2025,ryoo2023tokenturingmachines}, \cf~\Cref{fig:teaser} (right).  Their complexity is $O(1)$ \underline{at each step $t$} with respect to the number of input observations.\footnote{Here we ignore complexity \wrt. memory size, which is constant over time.}
observations (here we ignore complexity \wrt. memory size, which is constant over time).
This solves several issues by decoupling memory size from network capacity, allowing them to scale. In this work, we study the differences between \textit{Recurrent Transformers} \citep{moog2024,kinaema2025,ryoo2023tokenturingmachines} 
and their classical counterparts \citep{vaswani2017attention}, 
which for the sake of clarity we name \textit{History Transformers}, under the lens of their underlying \emph{compression mechanisms}.

\vspace*{0.5mm}
\textbf{Recurrent Transformers (RTs)} maintain some form of memory $\mathbf{M}_t$ and at each time step $t$ have access to the current observation $\mathbf{o}_t$ only. Memory is limited and they need to decide which information from $\mathbf{o}_t$ needs to be retained. Information not stored in $\mathbf{M}_t$ is lost forever.

\textbf{History Transformers (HTs)} do not have these shortcomings. At each time step $t$ they can access the full observation history and make use of it freely. Their drawback is the requirement to store the observation history and to attend to it.

\vspace*{0.5mm}
Both models must compress past observations into representations useful for future predictions. However, they do so in fundamentally different ways. HTs implicitly compress history when predictions are needed: attention dynamically selects relevant past tokens, and gradients can supervise all past representations directly. In contrast, RTs must explicitly decide at each time step what information to store in their fixed-size memory without having access to future requirements for prediction. Information not written to memory is irreversibly discarded, and supervision of these memory updates is indirect and delayed. As a result, learning effective memory representations becomes a significantly harder optimization problem.

We propose to bridge this gap by turning memory learning into a supervised compression task. We consider a History Transformer augmented with a fixed-size bottleneck representation that explicitly compresses its observation history, similar to the Perceiver Resampler architecture~\citep{perceiver}. We then distill this bottleneck representation into the memory of a Recurrent Transformer, directly aligning the student's memory with the teacher's learned compression. In this way, we transfer the compression strategy itself.
We validate our approach, called \textit{\ours}~for \textbf{C}ompressing \textbf{H}istory \textbf{I}nto \textbf{Me}mo\textbf{r}y with Tr\textbf{a}nsformers, primarily on Mem-RPE. This task provides a detailed study of long-horizon memory, including comparisons with existing recurrent models, generalization to longer sequences, ablations of the distillation procedure, and analysis of the learned representations. 
Distillation substantially narrows the gap to the full-history teacher while preserving the student's inference architecture and memory budget. 
The student can even outperform the teacher on long sequences.
Our approach is further validated on streaming VQA with a frozen LongVA-7B backbone~\citep{longva}.

\section{Related work}
\label{sec:related}

\myparagraph{Dealing with long context.} 
Causal models and KV-caching have decreased computational complexity, as the attention distributions for past tokens do not need to be recomputed at each time step.
Linear formulations of attention lead to significant speed-ups but lack expressivity \citep{lineartr2020}.
The \textit{Markovian Thinker} proceeds in chunks, each starting 
with a carry-over from the past history \citep{markovianthinker2026}.
This strategy has been applied to 3D reconstruction (\cf below).

\myparagraph{Recurrent models} initially had  vectorial hidden states \eg \textit{LSTM} \citep{DBLP:journals/neco/HochreiterS97} and \textit{GRU} \citep{cho-etal-2014-learning}.
Their memory scaling problem was addressed by external neural memory \citep{DBLP:journals/corr/GravesWD14,endtoendmemorynetworks2015,neuralgpu2016}. Recurrence is currently re-emerging, for instance through state space models like \textit{S4} \citep{s4iclr2022}, \textit{Mamba} \citep{gu2024mamba} and \textit{LRU} \citep{lru2023} inspired by control theory. \textit{xLSTM} adds QKV-attention for updates \citep{beck2024xlstm}, and full-fledged Transformers handling a memory bank are proposed by \textit{MooG} \citep{moog2024},  \textit{TTM} \citep{ryoo2023tokenturingmachines} and \textit{Kinaema} \citep{kinaema2025}. \textit{TRecViT} \citep{trecvit2026} runs RNNs over the embeddings of ViTs for video.

\myparagraph{Long-context in 3D reconstruction and robotics:} data driven reconstruction, pioneered by
\textit{DUSt3R} \citep{wang2024dust3r} and \textit{MASt3R} \citep{mast3r2024}, first integrated long context with global optimization but then evolved into applying transformers over the full sequence as in \textit{MUSt3R} \citep{must3r2025} and \textit{VGGT} \citep{vggt2025}. For robotics applications this was not considered sufficient as the base models could not deal with longer episodes. SLAM variants shifted to ``stitching'' based methods, where reconstruction is run on chunks and results are integrated. Examples are \textit{S-MUSt3R} \citep{smust3r2026}, \textit{VGGT-long} \citep{vggtlong2025} and \textit{VGGT-SLAM 2.0} \citep{vggtslam2_0}. Chunks have also been integrated end-to-end with TTT layers \citep{chen2026ttt3r,jin2026zipmap,xie2026scal3r,zhang2026loger}. \textit{CUT3R} \citep{cut3r2025} is a recurrent model for this task, which maintains a flat vectorial memory integrated into reconstruction with cross-attention. \textit{Kinaema} \citep{kinaema2025} is a recurrent model which maintains an implicit memory bank, avoids 3D reconstruction altogether and directly trains for memory-based map-free pose estimation. 

\myparagraph{Distillation} was initially proposed by \cite{distillinghinton2015} to transfer the power of larger teacher models into smaller students.
Distillation can also create universal models by distilling multiple teachers with different strengths or even tasks into a single model, as in \textit{DUNE} \citep{dune2025} or \textit{AM-RADIO} \citep{amradio2_5_2025}.
While smaller models are often inherently more efficient, distillation can also target students with specific architectures tailored for efficiency. In the context of robotic world models, \textit{DreamDojo} distills a bidirectional transformer into a causal one and with fewer diffusion steps \citep{dreamdojo2026}. Recent work has distilled transformers into state space models \citep{bick2024transformers,tr2mambawant2025}. Our formulation differs in that (1) the student is a recurrent transformer which features distributed memory (a memory bank), and (2) we distill the compression mechanism and therefore directly address the shortcoming of recurrent transformers.

\myparagraph{Connections between recurrent and history transformers:} recurrence has been added to classical transformers either through recurrent supervision in \textit{NextLat} \citep{nextlat} or through additional latent pathways in \textit{Full-Bandwidth Transformers} \citep{fullbandwidth}. Both keep context and target improved reasoning, but unlike ours they provide neither constant efficiency nor constant memory. \citet{kumar2026pretraining} supervise training of recurrent memory in single step predictions from memory generated with an encoder-decoder model which has seen context. In contrast, we distill over sequences and do not rely on passing dynamics information through subsequent gradient optimization steps, and we implement full-fledged systems targeting robotics and streaming computer vision.

\section{Distilling recurrent transformers}
\label{sub:method}

\myparagraph{Problem formulation.}
We consider tasks where a growing history of observations $\{\obs_{t'}\}_{t'\in1:t}$ is taken as input, and a prediction is required at a query time $t$. We particularly consider tasks where an individual observation $\obs_{t'}$, $t'<t$, can have a large impact on the prediction, including observations from far in the past. Predictions are conditioned on a query $\mathbf{q}_t$ that is distinct from the incoming observations. The query describes what information should be retrieved from the history: it is a goal image for Mem-RPE (``\textit{What is the pose of this image w.r.t. your current pose?}'') and a textual question for streaming VQA. We keep the method formulation general, then present our main application to Mem-RPE in \Cref{sec:memrpe} and an additional validation on streaming VQA in \Cref{sec:vqa}.

\begin{figure} \centering
    \resizebox{0.75\linewidth}{!}{
    \begin{tikzpicture}
    \draw (0, 0) node[anchor=west,inner sep=0] {
        \includegraphics[width=\linewidth]{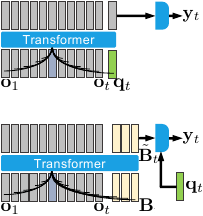}
    };
    \draw (0, 0.1) node[anchor=west,inner sep=0] {\textbf{(a) History Transformer}};
    \draw (0, -4.7) node[anchor=west,inner sep=0] {\textbf{(b) Latent Bottleneck History Transformer}};
    \end{tikzpicture}
    }
  \caption{\label{fig:bottleneck}\textbf{\textit{Latent Bottleneck History Transformer}} teacher (b). Compared to LLM-style models (a), sequence information   is compressed into a  fixed-size set of
  embeddings $\bottt_t$ before going through the prediction head conditioned on the query $\mathbf{q}_t$.}
\end{figure}

For \textit{History Transformers} this prediction is done using the full sequence  $\{ \obs_{t'} \}_{t'{\in}{1:t}}$ plus the query $\mathbf{q}_t$, 
\begin{equation}
\Big( \{\obs_1, \obs_2, \mydots, \obs_t\}\ ,\ \mathbf{q}_t \Big)  \trmappingl \mathbf{y}_t.
\end{equation}
Concretely, as shown in~\Cref{fig:bottleneck}(a), a set of multi-head self-attention blocks would process $\{ \obs_{t'} \}_{t'{\in}{1:t}}$ to obtain a set of token features which then, potentially combined with query features, would be fed to a prediction head.
In streaming and robotics settings, as the sequence length grows, the cost of both storing the sequence and attending to it becomes large and at some point prohibitive. 

In contrast, \textit{Recurrent Transformers}
maintain Memory $\mem_t$ and predict from it:
\begin{equation}
( \mem_{t-1}, \obs_t ) \trmappingl \mem_t, \quad
( \mem_t, \mathbf{q}_t )  \trmappingl \mathbf{y}_t.
\end{equation}
They do not require storing a history of observations and only access the current observation $\mathbf{o}_t$ and, more importantly, they do not need to \textit{attend} over an observation history.
In what follows, we will describe a method for distilling a History Transformer teacher into a Recurrent Transformer student. 

\myparagraph{Teacher: Latent Bottleneck History Transformer.} We will first motivate a variant of the History Transformer architecture which does not change its computational complexity, but rather introduces an explicit compression mechanism which, once trained, can be distilled into Recurrent Transformers. The ``\textit{Latent Bottleneck History Transformer}'' (LBHT), illustrated in~\Cref{fig:bottleneck}(b) will be used as teacher in the subsequent distillation process.

The LBHT adds a set of $B$ trainable read-out tokens $\bott=\{\bott^{(1)}, \bott^{(2)}, \mydots, \bott^{(B)} \}$ to the inputs and contextualizes them into output tokens $\bottt_t$. This bears resemblance to the \textit{Perceiver Resampler}~\citep{perceiver} or a similar mechanism in \textit{Flamingo}~\citep{flamingo}, and also to action read-out tokens in action experts of VLAs~\citep{pi05,smolvla2025}. 
The model learns how to take a full sequence of observations and compresses them into a memory bank in the form of the contextualized tokens $\tilde{\mathbf{B}}_t$. Most importantly, this compression does not have access to the query, \ie the goal of the task. The bottleneck $\bottt_t$ can therefore be interpreted as a form of latent representation, or latent ``map'', of the observations, which should be potentially useful for any task querying this content.

\myparagraph{Teacher: causal vs. non-causal}. Transformers come in two variants depending on how attention is computed. Transformers in modern LLMs are causal, \ie 
past tokens are not re-contextualized with information from newer tokens. Together with KV-caching and teacher forcing during training, this reduces complexity significantly, although 
the model still needs to attend to all past tokens for each prediction made at each time step $t$. In this work we opted for a non-causal transformer for three reasons: 
our experiments show that a non-causal model provides significant gains, which we explain with the advantages of bidirectional attention, making alignment more powerful.
Furthermore, we do not perform next token prediction, so teacher forcing during training does not apply. Finally, the teacher is not deployed, its computational complexity is thus of somewhat lesser importance. Note that even non-causal models do not ``\textit{cheat}'' in streaming applications, as the model does \textit{not} attend to ``future'' frames. Rather, non-causal models at time $t$ allow tokens from time $t{-}\tau$ to be re-contextualized by newer tokens from $t{-}\tau{+}1$ up to $t$, but not beyond $t$.

\myparagraph{Student: Recurrent Transformers.}
In recurrent models, observations are processed sequentially, and at each time step $t$ only the current observation $\mathbf{o}_t$ is available. The model maintains and updates a fixed-size memory $\mem_t$, \cf~\Cref{fig:teaser} (right), integrating new information at each step through the update equation:
\begin{equation}
\mem_t = f_\theta (\mem_{t-1}, \obs_t),
\end{equation}
where $f_\theta$ is a trainable function like an RNN, LSTM, GRU, or in our case a transformer as in \textit{Kinaema}~\citep{kinaema2025}. 
The interpretation of this recurrent memory $\mem_t$ is similar to the teacher's bottleneck embeddings $\bottt_t$: both are a latent representation of the observed history which should potentially be useful for any later task querying it.
However, its handling and maintenance by the model is fundamentally different and is based on a different compression mechanism. The recurrent model needs to decide which information to retain from the current observation $\obs_t$ at each time step, and all information not retained is lost for the future. Blindly storing the full observation is of course not possible, so a smart lossy compression mechanism needs to be learned.\footnote{Compression mechanisms in recurrent networks have been analyzed in detail by \cite{dong_predictive_2019} in terms of mutual information and the information bottleneck criterion \citep{bialek_predictability_2001}.}
We argue, and will show in the experiments, that learning such a mechanism is harder than training a transformer with access to full observation history, but that the latter can act as a teacher for the former.

\myparagraph{Distilling into Recurrent Transformers.}
We propose to distill the memory content of a Latent Bottleneck History Transformer into a Recurrent Transformer (RT) model. As a key design choice, we align the dimensions of the teacher's latent representation $\bottt_t$ and the student's memory representation $\mem_t$ by using an identical number of embeddings and embedding dimensions. This alignment enables direct distillation via an $L_1$ loss and in practice, we apply the $L_1$ loss between the flattened representations $\bottt_t$ and
$\mem_t$. 
Note that in case of dimension mismatch, 
a linear layer on $\mem_t$ could be added.
The distillation procedure is based on the following two key hypotheses:

\begin{figure*}[t] \centering
    \includegraphics[width=\linewidth]{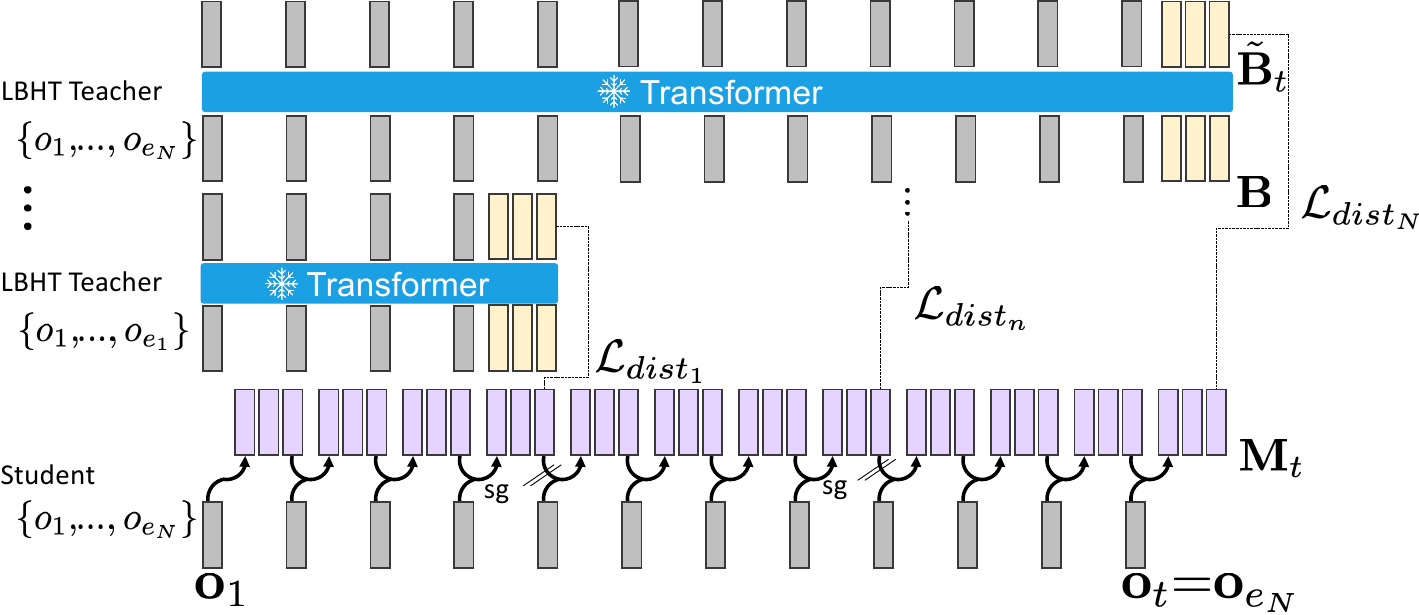} \\
    \caption{\label{fig:distillation}A student implemented as a recurrent transformer with \textit{Kinaema} architecture~\citep{kinaema2025} 
    is trained with distillation from a teacher implemented as \textit{Latent Bottleneck History Transformer (LBHT)}. A single rollout of full length of the student is aligned with multiple rollouts of the teacher over $N$ sub-sequences $\{\mathbf{o}_1,\mydots,\mathbf{o}_{e_n}\}_{n=1..N}$, 
    whose bottleneck embeddings $\bottt_{e_n}$ are distilled into recurrent memories $\mathbf{M}_{e_n}$ on corresponding time steps. Backpropagation of the student is limited to a fixed number of steps (shown as $sg$=``stop grad'') to ensure stability and limit GPU memory usage.}
\end{figure*}

\begin{description}[nosep,itemsep=0mm,labelindent=0mm,leftmargin=5mm,topsep=0mm]
    \item[H1: Learnability] ~\\ [-0.1cm]
\textrm{It is easier to learn to compress a full sub-sequence, \ie}
$\{\obs_1, \obs_2, \mydots, \obs_t \}$ $\trmapping \bottt_t,$
\\
\textrm{than it is to learn to compress it through recurrence, \ie} $( \mem_{t-1}, \obs_t )$ $\trmapping \mem_t.$
\item[H2: Expressivity of $\bottt_t$ through recurrence]\leavevmode \vspace{1mm} \\ [-0.1cm]
A sequence of bottleneck representations $\{\bottt_1, \bottt_2, \mydots, \bottt_t\}$ obtained by applying a trained LBHT teacher to subsequences of increasing lengths
can be sufficiently approximated through a learned recurrence function, \ie
$( \bottt_{t-1}, \obs_t ) \trmapping \bottt_t.$
\end{description}

\vspace*{1mm} \noindent
Both of these hypotheses will be verified experimentally, 
showing that History Transformers can be trained on substantially longer sequences than Recurrent Transformers, which suffer from stability issues when trained with downstream losses only. \textit{MooG}~\citep{moog2024} 
was trained on sequences of 8 steps, \textit{TRecViT}~\citep{trecvit2026}
for 64 steps and \textit{Kinaema}~\citep{kinaema2025}
for 100 steps, and we confirm this limitation in our own Mem-RPE experiments. In contrast, we could train HTs over visual sequences of up to 400 steps before running out of GPU memory and without encountering stability issues. 

\myparagraph{Segment-wise distillation.}
To best benefit from how the teacher model has learned a strategy to compress sequences of different lengths, we propose segment-wise distillation, illustrated in~\Cref{fig:distillation}. 
A rollout of the student over a sequence of length $t$ is decomposed into $N$ equal segments. Segments are indexed by $n$, their end time step by $e_n$ and the corresponding recurrent memory bank by $\mem_{e_n}$. We distill the memory bank at the end of each segment with the following $L_1$ loss:
\setlength{\abovedisplayskip}{3pt}
\setlength{\belowdisplayskip}{3pt}
\begin{equation}
\mathcal{L}_{dist} =
\sum_{n=1}^{N} \mathcal{L}_{dist_n} =
\sum_{n=1}^{N}
\left | \left |
\mem_{e_n} - \bottt_{e_n}
\right | \right |_1 \,.
\label{eq:distloss}
\end{equation}
As shown in~\Cref{fig:distillation}, a single forward pass over the student produces all the 
memory representations $\mem_{e_n}$, whereas each individual bottleneck representation $\bottt_{e_n}$ is obtained with its own separate forward pass over the teacher, as we distill from a non-causal teacher model.
Let us note that different loss terms $\mathcal{L}_{dist_n}$ are attached to different time steps of the full sequence. 
To limit GPU memory usage, we limit backpropagation of each loss term to a maximum allowable length. In the experimental section we experiment with different segment lengths and backpropagation horizons. When backpropagation horizon is larger than the segment length, gradient calculations ``overlap'', \ie for a given time step $e_n$, gradients from multiple loss terms are accumulated.

The teacher is first trained with the downstream task loss and then frozen for distillation. Student training combines downstream supervision with memory distillation:
$
\mathcal{L}_{student}=\mathcal{L}_{task}+\lambda_{dist}\mathcal{L}_{dist}.
\label{eq:studentloss}
$
The task loss and its supervision points are defined by the application. 


\section{Application to memory-based relative pose estimation}
\label{sec:memrpe}

\myparagraph{Task.} We evaluate our method on a challenging task in Embodied AI, which requires building a representation from visual observations collected along a trajectory and answering visual queries from it. Given a query image, the task is to predict the pose of the shown scene with respect to the robot's current pose. This task is related to map-free localization \citep{mapfreebrachmann2026} but, as introduced in \citep{kinaema2025} as ``\textit{Mem-RPE}'' (memory-based relative pose estimation), features additional challenges specific to Embodied AI: (1) observations are processed sequentially as sensed by an agent, and (2) the coordinate frame for pose estimation is centered on the 
agent, and is thus a \textit{moving coordinate frame}. This makes the task realistic and goal oriented, as an agent can take decisions based on its state. 
Agents move in 3D environments and at each time step $t$ receive a pair of sensor readings $\mathbf{o}_t=(\textbf{x}_t, \mathbf{u}_t)$, where  $\mathbf{x}_t \in \mathbb{R}^{3{\times}H{\times}W}$ is an RGB image of size $112{\times}112$, and $\mathbf{u}_t \in \mathbb{R}^7$ is an odometry estimate in the form of a difference of agent poses between $t$ and $t{-}1$. The task is conditioned on a
query image $\mathbf{q}_t \in \mathbb{R}^{3{\times}H{\times}W}$ and
consists in predicting pose $\mathbf{y}_t = (\mathbf{t},\mathbf{R})$, where $\mathbf{t} \in \mathbb{R}^3$ is the translation and
$\mathbf{R} \in SO(3)$ is the rotation matrix.

\myparagraph{Observation encoding.} Both student and teacher encode observations $\mathbf{o}_t = ( \textbf{x}_t, \mathbf{u}_t )$ in the same way: visual inputs are encoded via a fine-tuned DINO-v2 ViT-s~\citep{dinov22024}, where each of the 64 patch embeddings is down-projected to 64 values. Odometry $\mathbf{u}_t$ is embedded via an MLP into 64 values. The two representations are concatenated channel-wise into $\tilde{\mathbf{o}}_t = ( \tilde{\textbf{x}}_t, \tilde{\mathbf{u}}_t )$ of dimension 4160. 

\myparagraph{Student architecture.}
Our student is based on the \textit{Kinaema} model~\citep{kinaema2025}, a high-capacity recurrent architecture maintaining and updating a distributed memory state $\mem_t$ consisting of $B{=}20$ embeddings of dimension $E{=}3072$. More details are given in Appendix \ref{sec:studentarch}.

\myparagraph{Teacher architecture.}
We design a teacher tailored for distilling the \textit{Kinaema} student given above. The teacher is an LBHT (see~\Cref{sub:method} and~\Cref{fig:bottleneck}) with the same embedding size $E{=}3072$ and with $B{=}20$ bottleneck embeddings. By design, the bottleneck tensor of the teacher, $\bottt_t$, and the memory bank of the student, $\hmem_t$, have thus the same tensor shape. The teacher receives the same type of encoded observations $\tilde{\mathbf{o}}_t$, but observation encoders are not shared between teacher and student. 
The teacher has 7 transformer layers compared to the 3 layers of the  \textit{\seqmodname{}} student, while both architectures were 
optimized. The fact that the optimal number of layers is higher for the teacher than for the student is not necessarily surprising. It is well known that recurrent models tend to unroll some of their representation computations over time, additionally to performing them over layers.\footnote{As an extreme case one can consider a single-layer RNN, which has been shown to be Turing complete in spite of its simplicity~\citep{rnnsturingcomplete1995}. However, this might induce delays in outputs, and it has been shown that 1-layer RNNs are fundamentally equivalent to multi-layer RNNs with delays~\citep{rnnsdelayed2020}.}

\myparagraph{Teacher training.} The teacher is trained by directly supervising pose estimation as in~\citep{kinaema2025}, $\mathcal{L}_{RPE} = \sum\mathop{}_{\mkern-5mu i}
  \big(\ 
    |\mathbf{t}_i - \mathbf{t}_i^*| + |\mathbf{R}_i - \mathbf{R}_i^*|
  \ \big),
$ where $(\mathbf{t}_i,\mathbf{R}_i)$ and $(\mathbf{t}_i^*,\mathbf{R}_i^*)$ are predicted and GT pose for training image $i$, respectively. The teacher also uses an auxiliary masked image modeling loss, which reconstructs the same query images after they have been masked through a second decoder head.
Training the student combines the distillation loss $\mathcal{L}_{dist}$ (Equation~\ref{eq:distloss}), with the downstream loss $\mathcal{L}_{RPE}$. 
The teacher is trained on sequences of randomized length of $50{\leq}T{\leq}400$, far beyond the limit of the original \textit{\seqmodname} model ($T{\leq}100$).

\myparagraph{Student distillation.} Masked image modeling is not used for the student. Distillation is done on sequences of randomized length $50{\leq}T{\leq}200$ with multiple segments (see \Cref{sub:method}), their numbers are ablated in the experimental section. 
The downstream loss $\mathcal{L}_{RPE}$ is additionally applied on each segment $n$ by taking the memory $\hmem_{e_n}$, predicting relative pose for $2e_n$ query images of two different types: (i) the $e_n$ observed images $\{ \mathbf{x}_t \}_{t\in 1:e_n}$, and (ii) $e_n$ \textit{alternative} images $\{ \mathbf{x}^{alt}_t \}_{t\in 1:e_n}$, which have \textit{not} been observed but lie in the previously observed region of the scene, generated in the simulator.

\begin{table}[t] \centering
    \caption{
        \textbf{Comparisons with the state of the art} on \rpetest.
    } 
    \label{tab:sota}
    \vspace{-0.1cm}
\resizebox{\linewidth}{!}
{\small
    \setlength{\tabcolsep}{1pt}    
    \begin{tabular}{lcc|C{T200T}C{T200T}C{T200T}|C{T800T}C{T800T}C{T800T}}
            \specialrule{1pt}{0pt}{0pt}
            \rowcolor{TableGray2}
             \cellcolor{TableGray1}\textbf{Model} &
             \textbf{{Obs}} &
             \multicolumn{1}{c}{\textbf{\small Mem}} &
             \multicolumn{3}{c|}{\textbf{Seq len 200}} &
             \multicolumn{3}{c}{\textbf{Seq len 800}}
             \\
             \rowcolor{TableGray2}
             \cellcolor{TableGray1}&
             \cellcolor{TableGray2}{\textbf{hist}} &
             \multicolumn{1}{c}{\cellcolor{TableGray2}\textbf{size}} &
             \metrics &
             \metrics
             \\[1pt]
            \specialrule{0.5pt}{0pt}{0pt}
            \cellcolor{TableGray1} \textbf{LBHT (Teacher)} &
            \cellcolor{TableGray2} \yes &
            \cellcolor{TableGray2} 61.4k &
            \textbf{44}&\textbf{56}&\textbf{72}&
            \textbf{16}&\textbf{22}&\textbf{44}
            \\            
            \specialrule{0.5pt}{0pt}{0pt}
            \cellcolor{TableGray1} \textbf{MooG} \citep{moog2024} &
            \cellcolor{TableGray2} \no &
            \cellcolor{TableGray2} 524.3k &
            0&5&14&
            0&3&9
            \\
            \cellcolor{TableGray1} \textbf{LRU}  \citep{lru2023} &
            \cellcolor{TableGray2} \no &
            \cellcolor{TableGray2} {3.1k} &
            4&18&34&
            2&9&20
            \\
            \cellcolor{TableGray1} \textbf{EMA} \citep{ema2025} &
            \cellcolor{TableGray2} \no &
            \cellcolor{TableGray2}153.6k&
            6&18&34&
            3&11&24
            \\
            \cellcolor{TableGray1} \textbf{xLSTM} \citep{beck2024xlstm} &
            \cellcolor{TableGray2} \no &
            \cellcolor{TableGray2} 2,359.3k&
            8&23&47&
            5&13&29
            \\
            \cellcolor{TableGray1} \textbf{GRU}  \citep{cho-etal-2014-learning} &
            \cellcolor{TableGray2} \no &
            \cellcolor{TableGray2} {3.1k} &
            12&32&56&
            4&14&31
            \\
            \cellcolor{TableGray1} \textbf{Kinaema} \citep{kinaema2025} &
            \cellcolor{TableGray2} \no &
            \cellcolor{TableGray2} 61.4k &
            21&41&63&
            10&21&37
            \\
            \cellcolor{TableGray1} \textbf{\ours} (ours)  &
            \cellcolor{TableGray2} \no &
            \cellcolor{TableGray2} 61.4k &
            \textbf{38}&\textbf{45}&\textbf{65}&
            \textbf{22}&\textbf{26}&\textbf{48}
            \\
            \specialrule{1pt}{0pt}{0pt}
    \end{tabular}   
}
\end{table}

\myparagraph{Experimental setup.}
We use the data and protocol from~\cite{kinaema2025}\footnote{\url{https://github.com/naver/kinaema}}: data was generated in the Habitat simulator~\citep{habitat2} from the HM3D \citep{ramakrishnan2021hm3d} and Gibson \citep{xia2018gibson} datasets with 
independent scene splits related to Mem-RPE, namely \rpetrain: 800 scenes from the HM3D/train split;  \rpeval: 72 scenes from the Gibson/train split; \rpetest: 100 scenes from the HM3D/val split. All result tables have color-coded backgrounds indicating the splits. 
Actions are chosen from a discrete action space, \eg $\mathcal{A}$ ={\small \tt \{move forward, turn left, turn right\}}. The test set shows some out-of-distribution settings as (1) test sequences are significantly longer than the training sequence length (up to $T{=}800$); (2) Motion between subsequent poses is different, as forward is 25cm instead of 10cm, and turning angles are 10$^\circ$ instead of 5$^\circ$.

\myparagraph{Metrics.} We report percentage of correctly estimated poses with three thresholds as in~\citep{kinaema2025}: $<1m$ of translation and $<10^\circ$ of rotation errors, $<1$m and $90^\circ$, and $<2$m and $90^\circ$. 

\myparagraph{Comparison to the state of the art.}
~\Cref{tab:sota} provides a comparison with the state of the art in terms of Mem-RPE accuracies on \rpetest. The LBHT teacher is the only model with access to the observation history and therefore is not comparable to the other models. All other models are recurrent and we report them with their  corresponding memory size.
\textit{\ours}{} significantly outperforms the state of the art, including \textit{Kinaema} 
that shares the exact same architecture, including all hyper-parameters (numbers of layers, heads, embedding size etc.). This shows the effectiveness of our proposed compression strategy to distill from a LBHT teacher.
On sequence length 200, our method performs  close to the teacher, with \eg 65\% accuracy at 2m90\textdegree  \vs 72\% for LBHT and 63\% for \textit{Kinaema}.
Interestingly, \textit{\ours{}} even performs slightly better than LBHT on long sequences and significantly better than Kinaema. 
All remaining ablations and analysis below have been performed on \rpeval.

\begin{figure}[t]
    \centering
    \begin{minipage}[t]{0.49\linewidth}
        \vspace{0pt}
        \centering
        \includegraphics[width=\linewidth]{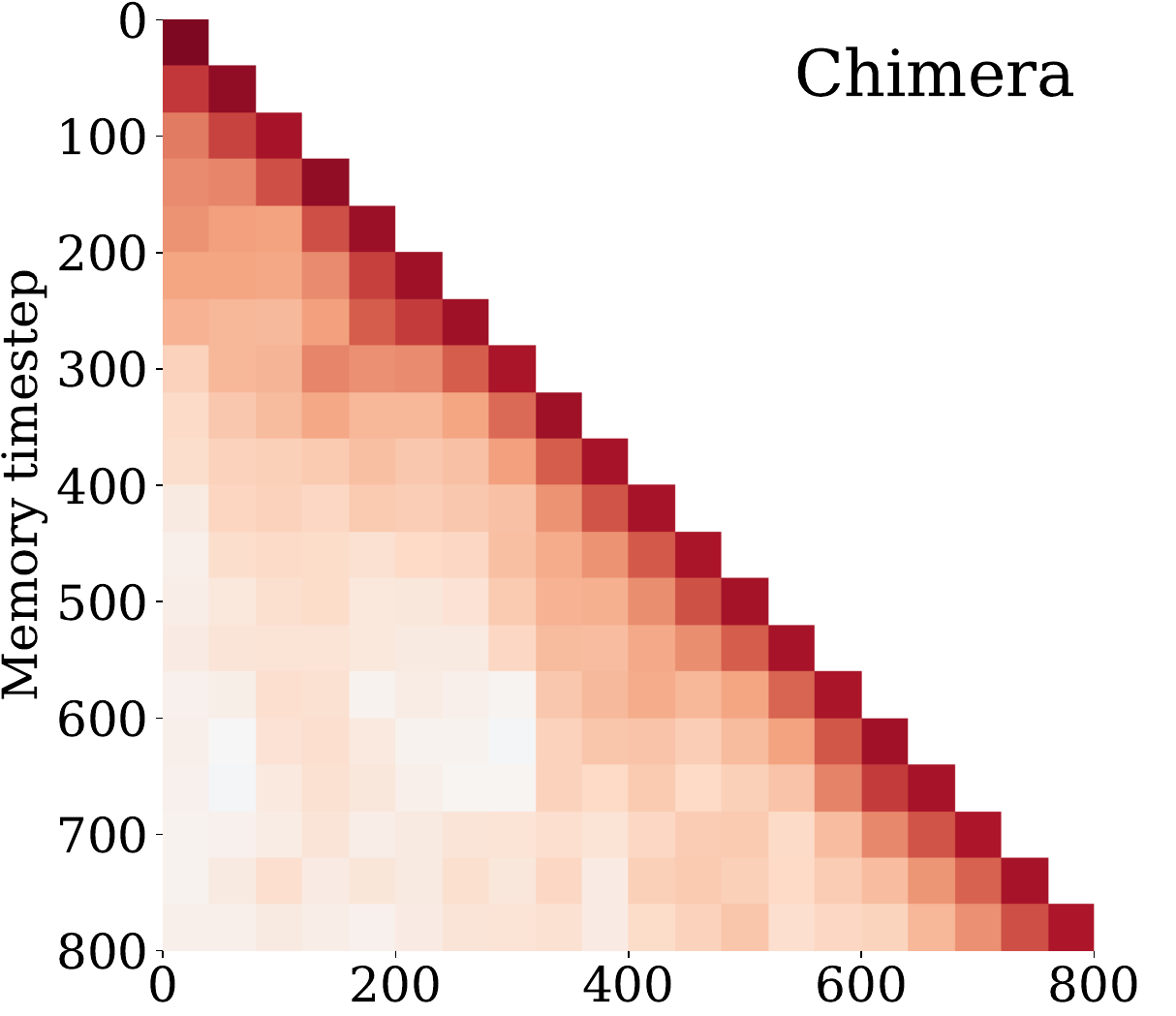}
    \end{minipage}
    \hfill
    \begin{minipage}[t]{0.49\linewidth}
        \vspace{0pt}
        \centering
        \includegraphics[width=\linewidth]{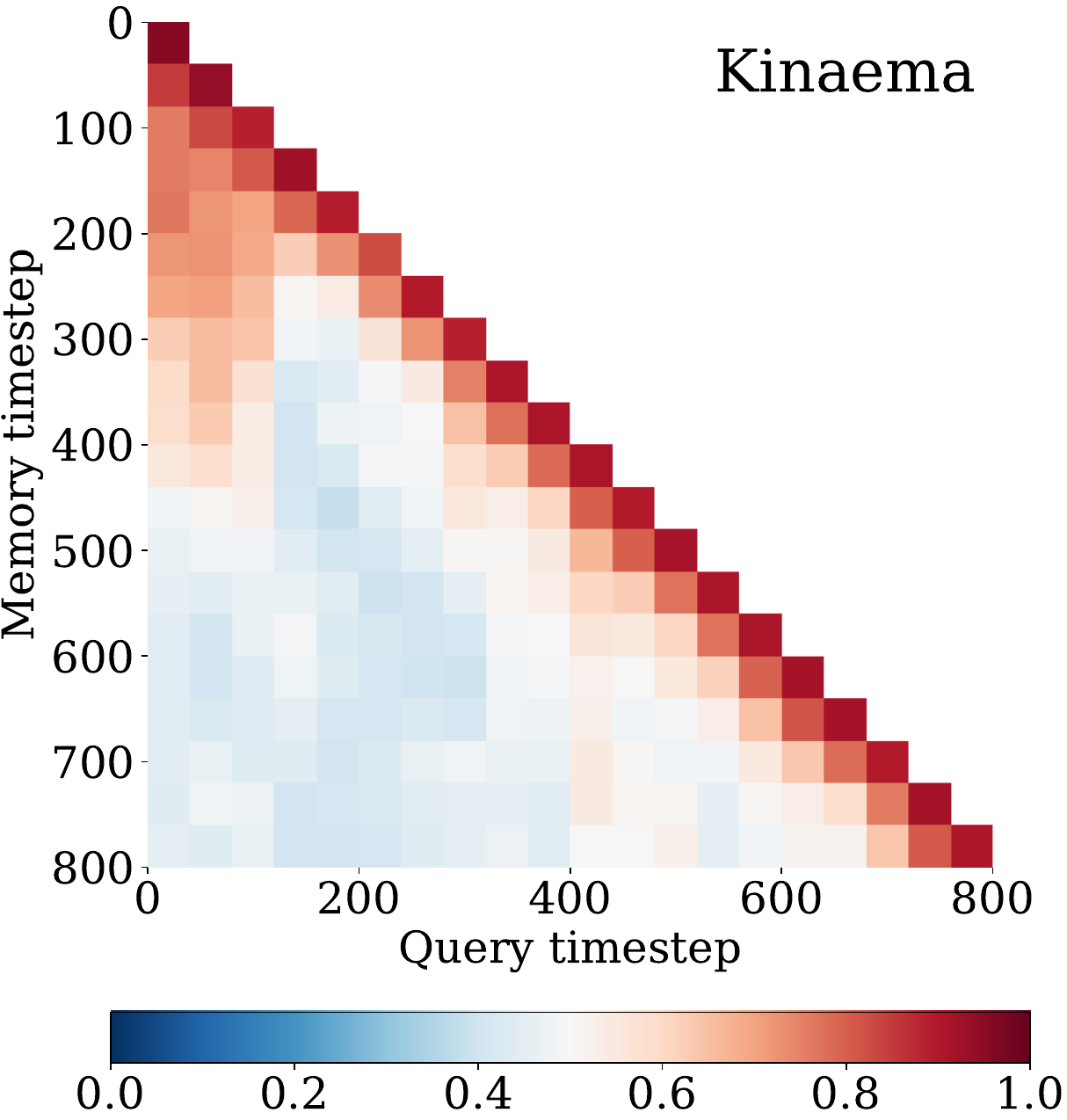}
    \end{minipage}
    \caption{\textbf{Comparison between \textit{Chimera} and \textit{Kinaema}} in terms of Mem-RPE performance (2m 90\textdegree) for varying query and memory timesteps. Each cell $(t,t')$ corresponds to accuracy for queries $[\query_{t'-40},\query_{t'}]$ from memory $\mem_{t}$.}
    \label{fig:rpe_matrix}
\end{figure}

\begin{figure*}[t]
    \centering
    \includegraphics[width=0.44\linewidth]{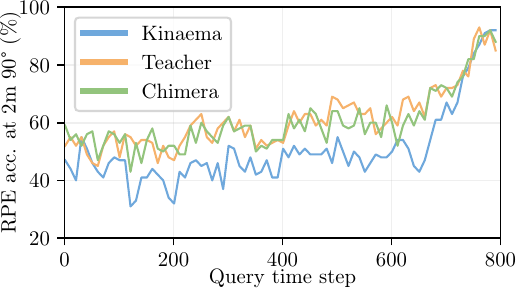}
    \hfill
    \includegraphics[width=0.44\linewidth]{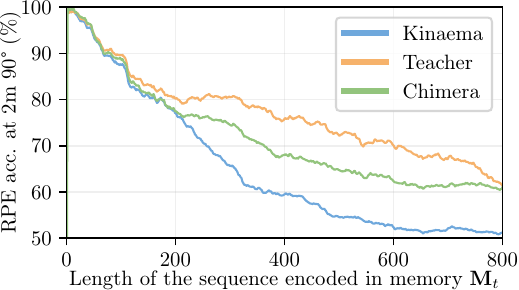}
    \caption{\label{fig:rpe_vs_t}\textbf{Mem-RPE performance as a function of query age and encoded sequence length} for the teacher, our model, and \textit{Kinaema} on \rpeval (accuracy at 2m 90\textdegree).
    \emph{Left:} We vary $t$ on the x-axis and consider memory $\hmem_{800}$ having encoded a full sequence of 800 frames. \emph{Right:} we vary the length $t$ of the sequence encoded into memory $\hmem_t$ and query it with frames $\leq t$. All queries are ``alternative'' frames, \ie close to observations in the scene, but \textit{not} actually observed.} 
\end{figure*}

\myparagraph{Mem-RPE performance with varying query age.}
\Cref{fig:rpe_vs_t} (left) shows how performance evolves when memory $\bf{M}_{800}$ is fixed after 800 observed frames, and considering queries further in the past, \ie earlier in the sequence, or close to $t{=}800$ and 
thus to the most recent observation $\obs_t$ for the memory. Note that we do not query observed frames but \textit{alternative} images, \ie query images close to actual observations. Performance increases on the right part of the plots for all methods as the queries become closer to the last observations encoded within the memory. We observe that \textit{\ours}{} performs better than \textit{Kinaema}
and close to the LBHT teacher at all timesteps. 

\myparagraph{Mem-RPE performance with varying eval sequence length $T$.}
In~\Cref{fig:rpe_vs_t} (right), we show how Mem-RPE accuracies evolve when considering subsequences of various lengths, up to 800. \textit{\ours}{} outperforms \textit{Kinaema} for all sequence lengths. 
Interestingly, the LBHT teacher performance drops steadily for sequences longer than 400 frames, used for its training, while our recurrent model is more stable for longer sequences despite being trained with sequences of at most 200 frames. 
\Cref{fig:rpe_matrix} further compares our approach to \textit{Kinaema} by considering all ranges of queries and memories timesteps. The diagonal corresponds to queries that are close to the current position, explaining its high performance. 
We observe overall that Chimera significantly outperforms Kinaema, in particular for large values of memory timesteps.

\myparagraph{Impact of model choice when full-scale trained.}
\Cref{tab:teachertypes} compares several base models in terms of Mem-RPE accuracies on \rpeval when training on long sequences of up to $T{=}400$ and for 600 epochs. We compare HT without bottleneck, \cf~\Cref{fig:bottleneck}(a), with the LBHT teacher, \cf~\Cref{fig:bottleneck}(b). In this large-scale training setting, \textit{Kinaema} diverges; the baseline transformer (a) obtains excellent accuracy for sequence length 200, that is among the sequence 
lengths seen during training. However, it performs poorly on longer sequences, which might come from the difficulty of applying transformers on sequences significantly longer than the ones seen during 
training~\citep{zhao2024length,huang2024formal}. In contrast, LBHT is significantly less affected by this phenomenon. 

\begin{table}[t] \centering
    \caption{\label{tab:teachertypes}\textbf{Impact of model choice on full-scale training} (seq. length $T{=}400$ and full 600 epochs). \textit{Kinaema} diverges when trained on these scales (\rpeval).}
    \vspace{-0.1cm}
    \resizebox{\linewidth}{!}{
\begin{tabular}{l|C{T200}C{T200}C{T200}|C{T800}C{T800}C{T800}}
            \specialrule{1pt}{0pt}{0pt}
            \rowcolor{TableGray2}
             \cellcolor{TableGray1}\textbf{Model} &
             \multicolumn{3}{c|}{\textbf{Seq len 200}} &
             \multicolumn{3}{c}{\textbf{Seq len 800}}
             \\
             \rowcolor{TableGray2}
             \cellcolor{TableGray1} \textbf{}&
             \metrics &
             \metrics
             \\[1pt]
             \specialrule{0.5pt}{0pt}{0pt}
             \cellcolor{TableGray1} \textbf{History Transformer}&
             \textbf{59}	& \textbf{74}	& \textbf{85}	& 6	& 19	& 35
             \\
             \cellcolor{TableGray1} \textbf{Latent Bottleneck History Transformer}&             
              28	& 51	& 74	& \textbf{16}	& \textbf{31}	& \textbf{55}
             \\
             \cellcolor{TableGray1} \textbf{Kinaema} \citep{kinaema2025} &
             \multicolumn{3}{c|}{\cellcolor{T200}diverges}&
             \multicolumn{3}{c}{\cellcolor{T800}diverges}
             \\
            \specialrule{1pt}{0pt}{0pt}
    \end{tabular}
    }
\end{table}

\myparagraph{Design choices for LBHT teacher training.}
\Cref{tab:ablationsteacher} 
reports a sensitivity analysis of varying the maximum length $T$ of training sequences and the number of epochs when training the LBHT teacher.
Large maximum sequence lengths (\eg 400) are necessary, and longer 
sequences require longer training too. Causal transformers, where past tokens are not re-contextualized by newer observations within the available prefix, perform substantially worse on this task.

\begin{table}[t] \centering
    \caption{\label{tab:ablationsteacher}\textbf{Design choices for LBHT teachers.} Top: different training sequence lengths $T$ and different training lengths for a non-causal model. Bottom: Impact of transformer causality $\mathcal{C}$ (\rpeval).}
    \vspace{-0.1cm}
    \centering
        {\small
        \setlength{\tabcolsep}{2pt}
        \begin{tabular}{cc|C{T200}C{T200}C{T200}|C{T800}C{T800}C{T800}}
        \specialrule{1pt}{0pt}{0pt}
        \rowcolor{TableGray2}
         \cellcolor{TableGray2}\textbf{\scriptsize Train} &
         \cellcolor{TableGray2} \textbf{\scriptsize Num}&
         \multicolumn{3}{c|}{\textbf{Seq len 200}} &
         \multicolumn{3}{c}{\textbf{Seq len 800}}
         \\
         \rowcolor{TableGray2}
         \cellcolor{TableGray2} \textbf{\scriptsize $T$}&
         \cellcolor{TableGray2} \textbf{\scriptsize epochs}&         
         \metrics &
         \metrics
         \\[1pt]
         \specialrule{0.5pt}{0pt}{0pt}
         \cellcolor{TableGray2} \textbf{400}&
         \cellcolor{TableGray2} \textbf{600}&         
         28	& 51	& 74	& 16	& 31	& 55
         \\
         \cellcolor{TableGray2} \textbf{400}&
         \cellcolor{TableGray2} \textbf{300}&
         21	& 49	& 73	& 10	& 26	& 50
         \\
         \cellcolor{TableGray2} \textbf{400}&
         \cellcolor{TableGray2} \textbf{200}&         
         9 & 	28 & 	51	& 6	& 19	& 40
         \\
         \cellcolor{TableGray2} \textbf{300}&
         \cellcolor{TableGray2} \textbf{200}&
         12	& 34 &	59 &	6 & 	19 &	42
         \\
         \cellcolor{TableGray2} \textbf{200}&
         \cellcolor{TableGray2} \textbf{200}&
         16	& 40 &	65 &	6 &	18 & 	40
         \\
        \specialrule{1pt}{0pt}{0pt}
        \\
    \end{tabular}
    }    
        \setlength{\tabcolsep}{1pt}
        \begin{tabular}{ccc|C{T200}C{T200}C{T200}|C{T800}C{T800}C{T800}}
        \specialrule{1pt}{0pt}{0pt}
        \rowcolor{TableGray2}
         \cellcolor{TableGray2}\textbf{\scriptsize Train} &
         \cellcolor{TableGray2} \textbf{\scriptsize Num}&
         \cellcolor{TableGray2} $\mathcal{C}$&
         \multicolumn{3}{c|}{\textbf{Seq len 200}} &
         \multicolumn{3}{c}{\textbf{Seq len 800}}
         \\
         \rowcolor{TableGray2}
         \cellcolor{TableGray2} \textbf{\scriptsize $T$}&
         \cellcolor{TableGray2} \textbf{\scriptsize epochs}&
         \cellcolor{TableGray2} \textbf{}&
         \metrics &
         \metrics
         \\[1pt]
         \specialrule{0.5pt}{0pt}{0pt}         
         \cellcolor{TableGray2} \textbf{100}&
         \cellcolor{TableGray2} \textbf{200}&
         \cellcolor{TableGray2} \textbf{\no}&
         17& 	35	& 55	& 4	& 10& 	30
         \\
         \cellcolor{TableGray2} \textbf{100}&
         \cellcolor{TableGray2} \textbf{200}&
         \cellcolor{TableGray2} \textbf{\yes}&
         8	& 18	& 35	& 2& 	6&	16
         \\
         \cellcolor{TableGray2} \textbf{200}&
         \cellcolor{TableGray2} \textbf{300}&
         \cellcolor{TableGray2} \textbf{\no}&
         28	& 54& 	76	& 6	& 14& 	38
         \\
         \cellcolor{TableGray2} \textbf{200}&
         \cellcolor{TableGray2} \textbf{300}&
         \cellcolor{TableGray2} \textbf{\yes}&
         10	 & 22& 	40	& 4	& 10& 	27
         \\
        \specialrule{1pt}{0pt}{0pt}
        \\
        \end{tabular}
\end{table}

Beyond task performance, we analyze how distillation affects the internal structure of the learned memory.
We compare the memory representations of the baseline \textit{Kinaema} model and \textit{\ours}{} along two axes:
(i) stability of memory over time;
(ii) the degree of specialization among individual memory tokens.

\myparagraph{Memory stability.}
\Cref{fig:mem_vs_t} shows the stability of the memory as the \textit{normalized memory update norm} $\mathcal{N}$, defined as the L1 norm of the memory update between two consecutive timesteps, normalized by the average norm of the memory over the sequences:
\begin{figure}[t]
    \centering
    \includegraphics[width=0.9\linewidth]{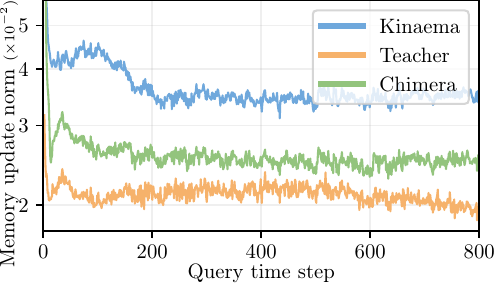}
    \caption{\label{fig:mem_vs_t}\textbf{Stability of memory} for the teacher, our model, and \textit{Kinaema} on \rpeval. Memory stability is measured by the update norm $\mathcal{N}$.} 
\end{figure}

\begin{figure}
    \centering
    \includegraphics[width=\linewidth]{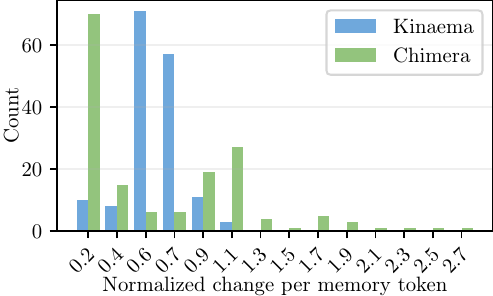}
    \caption{
    \textbf{Histogram of memory token changes} accumulated over a trajectory. Aggregated results  
    for \rpeval sequences of length $T=200$.\label{fig:mem_change_hist}
    }
\end{figure}
$
     \mathcal{N}_t = \frac{1}{\eta} || {\bf M}_t - {\bf M}_{t-1} ||_1 ,
     \label{eq:memnorm}
$
where $\eta  = \frac{1}{T}\sum_{t=1}^T || {\bf M}_t ||_1$.
\textit{\seqmodname}{} shows a higher rate of change up to $T{=}100$, for which it was trained, followed by a drop, which we conjecture to be linked to a lack of generalization of its memory management system: when new content is added to memory filled beyond amounts seen in training, it seems to change its update behavior. 
In contrast, the LBHT teacher exhibits a flat change rate over time and also beyond the length of $T{=}400$ on which it was trained, with an overall lower norm. 
The distillation process makes \textit{\ours}{}'s behavior similar to the teacher.

\myparagraph{Specialization of memory tokens.}
We investigate whether individual memory tokens develop specialized roles.
We compute the absolute change of each memory token at each time step, sum it across the trajectory, and then normalize by the total change across all tokens.
A uniform distribution would indicate that all tokens are updated equally, while a skewed distribution would suggest that certain tokens specialize for storing persistent information while others are more dynamic and frequently rewritten.
\Cref{fig:mem_change_hist} shows the results obtained on \rpeval with sequences of length $T=200$.
Both models exhibit a non-uniform distribution, suggesting that memory tokens do specialize.
Interestingly, the distilled model displays a more skewed distribution, with the majority of the tokens changing only a little while a few tokens account for a large portion of the change.
This suggests that \textit{\ours}{} learns to maintain a stable core of memory tokens that are updated less frequently, while capturing discriminative information in a few highly dynamic tokens.

\section{Additional application: streaming visual question answering}
\label{sec:vqa}

\begin{table}[t]
{ \small
    \begin{center}    
    \noindent
    \caption{\label{tab:vqa}\textbf{Streaming VQA results} (accuracies in \%).}

    \setlength{\tabcolsep}{3pt}
    \begin{tabular}{l|cc}
            \specialrule{1pt}{0pt}{0pt}
            \rowcolor{TableGray2}
             \cellcolor{TableGray1}\textbf{Model} &
             \textbf{OVO-} &
             \textbf{StreamingBench}
             \\
             \rowcolor{TableGray2}
             \cellcolor{TableGray1} &
             \textbf{Bench} &
             \textbf{(Real)}
             \\[1pt]
            \specialrule{0.5pt}{0pt}{0pt}
            \cellcolor{TableGray1} \textbf{No vision} & 33.5 & 39.3 \\
            \cellcolor{TableGray1} \textbf{Bottleneck teacher} & \textbf{37.0} & 49.4 \\
            \cellcolor{TableGray1} \textbf{Student w/o distillation} & 34.3 & 48.1 \\
            \cellcolor{TableGray1} \textbf{Student w/ distillation} & 36.8 & \textbf{49.8} \\
            \specialrule{1pt}{0pt}{0pt}
    \end{tabular}
    \end{center}
}
\end{table}
To complement our main evaluation on Mem-RPE, we apply bottleneck distillation to streaming visual question answering, where a model answers textual questions using the video observations available at query time. This experiment tests whether supervising the compressed history also benefits recurrent predictions in a video-language setting. We use LongVA-7B~\citep{longva} with a frozen visual encoder and language model, and train temporal compression modules on LLaVA-Video~\citep{llavavideo}. The LBHT teacher jointly compresses the observed frame representations into a bottleneck, while the recurrent student updates a fixed-size memory sequentially. Both provide visual context to the same language model. We compare students trained with the answer-prediction loss alone and with the additional segment-wise bottleneck supervision of \Cref{sub:method}. The appendix provides more details about the architecture, datasets, training setup, and teacher/student ablations.

Table~\ref{tab:vqa} reports accuracy on OVO-Bench~\citep{ovobench} and the `Real' task of StreamingBench~\citep{streambench}. Distillation improves the recurrent student from $34.3\%$ to $36.8\%$ on OVO-Bench and from $48.1\%$ to $49.8\%$ on StreamingBench Real, gains of $2.5$ and $1.7$ percentage points, respectively. The distilled student approaches the teacher on OVO-Bench and slightly exceeds its reported Real score. These gains provide additional evidence that our proposed memory distillation can also improve recurrent memory learning for video-language prediction.

\section{Conclusion}

We addressed the fundamental performance gap between History Transformers and Recurrent Transformers in long-horizon robotic 
visual
tasks and  argued that it is linked to the optimization difficulty associated with learning ``on-the-fly'' compression. We demonstrated that a full-history teacher's compression strategy can be distilled into the fixed-size memory of a recurrent student. 
Our approach uses an explicit teacher bottleneck to supervise the student's memory at multiple points along the observation sequence.
Our experiments on sequential map-free pose estimation and streaming visual question answering confirm Our experiments on sequential map-free pose estimation confirm
our two primary hypotheses: first, that supervising a student's memory with a teacher's bottleneck representation significantly eases the learning of complex temporal dependencies; and second, that recurrent architectures are sufficiently expressive to approximate these non-causal compression strategies. The resulting \textit{\ours{}} model retains the favorable $O(1)$ inference complexity of recurrent models while achieving performance levels previously reserved for computationally expensive transformers. As an additional application, streaming visual question answering with a frozen LongVA-7B backbone shows improvements on OVO-Bench and StreamingBench Real, providing further evidence for the training principle beyond pose estimation.


\small
\bibliographystyle{ieeenat_fullname}
\bibliography{main}

\renewcommand{\thesection}{\Alph{section}}
\renewcommand{\thefigure}{\Alph{figure}}
\renewcommand{\thetable}{\Alph{table}}
\setcounter{figure}{0}
\setcounter{table}{0}

\newpage
\appendix

\begin{center}
 \Large \textbf{Appendix}
\end{center}

\section{Mem-RPE architecture details}
\label{app:memrpe:architecture}

\subsection{Input encoders}

Input images are of resolution $112{\times}112$ and are encoded with a DINO-v2 \citep{dinov22024} ViT-s backbone that is initialized with DINO-v2 and finetuned. The encoder has a patch size $14{\times}14$, leading to $8{\times}8$ patch tensors per image. Patch embeddings are linearly down-projected to $64$ values, resulting in visual embeddings of dimension $4096$, which are combined with delta pose embeddings of size $64$.

\subsection{Student}
\label{sec:studentarch}

The student has the same architecture as \textit{Kinaema}~\citep{kinaema2025}. The model performs two operations, a recurrent memory update step, and a decoding step which queries memory given a query $\mathbf{q}_t$,
\begin{equation}
\hmem_t = \Update \big(\hmem_{t-1}, \tilde{\mathbf{o}}_t\big), 
\quad    
\mathbf{y}_t = \text{Decoder}\big(\hmem_t, \Encodegoal\mathbf{q}_t\big).
\end{equation}
The decoder is implemented as a transformer using cross-attention between the goal query and the memory embeddings $\mathbf{M}_t$.

The core innovation of \textit{\seqmodname}{} lies in its scalable and gated update mechanism. To handle high-capacity memory without  parameter growth, $\Update$ is decomposed into three stages:
\begin{enumerate}[nosep,itemsep=2mm,labelindent=0mm,leftmargin=5mm,topsep=2mm]
    \item \textbf{Correction:} each of the $B$ memory embeddings $\{\hmem_{t-1,b}\}_{b{\in}1:B}$ is channel-wise concatenated with the current encoded observations $\tilde{\mathbf{o}}_t$ and summed with a positional embedding $\mathbf{e}_b$, then projected: $$\hmem_{t,b}^{corr} = \Linear([\hmem_{t-1,b} + \mathbf{e}_b, \tilde{\mathbf{x}}_t, \tilde{\mathbf{u}}_t]).$$
    \item \textbf{Contextualization:}
    a transformer with self-attention  $\tilde{\hmem}_t = \SA(\hmem_t^{corr})$ allows for interaction between the $B$ distributed memory slots.
    \item \textbf{Shared Gating:} to ensure stability and model varying temporal dynamics, a GRU-based \citep{cho-etal-2014-learning} gating block is applied: $\hmem_{t,b} = \GRU(\hmem_{t-1,b}, \tilde{\hmem}_{t,b})$. Crucially, the gating weights are shared across all $B$ memory embeddings, decoupling the memory capacity from the gating complexity.
\end{enumerate}

Our student is identical, with the exception of \textit{Kinaema}'s masked image modeling head, not used here. The architecture details are:
\begin{itemize}
[nosep,itemsep=2mm,labelindent=0mm,leftmargin=5mm,topsep=2mm]
    \item The transformer block has 3 layers, 24 heads, and an MLP factor of 4. The gating block is a GRU with 3 layers.   
    \item \textbf{Memory reshape}. As in \textit{Kinaema} \citep{kinaema2025}, the $20$ tokens of dim $3072$ are reshaped into $160$ tokens of dim $384$ before passing them to the decoder. This allows the memory model to handle more values (higher dim) in the feed-forward layer of each memory token while at the same time allowing the decoder to compute fine-grained attention between query image patches and a larger number of memory tokens.
    \item The decoder (which queries into the updated memory to provide pose predictions) employs a cross-attention (CA) layer with no residual connection. Its keys and values are provided by the (reshaped) memory tokens and the queries are the patch tokens from the goal image encoder $\Encodegoal$. The motivation behind this design with missing skip connections is to guide the decoder on using the memory outputs by removing the direct residual path from query features to the prediction head. A learnable token (\ie a \tcls token) is attached to the output of the CA layer and given as input to a sequence of 4 standard self-attention (SA) blocks~\citep{vaswani2017attention}. Finally, the relative pose of the goal image is estimated by an MLP with 1 hidden layer from the \tcls  token.
\end{itemize}

\subsection{Latent Bottleneck History Transformer Teacher}

The transformer block of the Latent Bottleneck History Transformer (LBHT) teacher has 7 layers, 24 heads, and an MLP ratio of 4.0. The input embeddings of size $4160$ are linearly down-projected to size $E{=}3072$, and then token-wise concatenated with the $B{=}20$ bottleneck embeddings of the same embedding dim. The contextualized bottleneck embeddings $\bottt_t$ are used for distillation. For the RPE loss, the teacher uses the same decoder architecture as the student.

\section{Mem-RPE training details}
\label{app:memrpe:training}

\subsection{Training the LBHT teacher}

\resizebox{\linewidth}{!}{
\begin{tabular}{l|l}
\hline
 Batch size          & 32 \\
 Optimizer           & AdamW ($\beta_1=0.9, \beta_2=0.99, \epsilon=1e{-}15$)\\
 Base Learning rate  & 1.5e-4 \\
 Minimum LR          & 1e-8 \\
 Scheduler           & Cosine annealing \\
 Warm-up epochs      & 40 \\
 Gradient clipping   & 1.0 \\
 Weight decay        & 0.05 \\
 Epochs              & 600 \\
 Weight for RPE loss & 1.0 \\
 Weight for MIM loss & 1.0 \\
\hline
\end{tabular}
}

\subsection{Distilling the student}

To compute the distillation loss (\cref{eq:distloss}), 
we first flatten the memory representation, resulting in vectors of dimension 61,440. We then compute the sum of absolute differences. The hyperparameter $\lambda_{dist}$ that combines the loss $\mathcal{L}_{dist}$ with $\mathcal{L}_{RPE}$ 
is set to $3 \times 10^{-5}$ in all Mem-RPE experiments (Section~\ref{sec:memrpe}), \ie the training loss for a given segment $n$ is $\mathcal{L}_n = \lambda_{dist} \mathcal{L}_{dist_n} + \mathcal{L}_{RPE_n}$. 

Table~\ref{tab:wloss} provides an ablation on this hyperparameter. 
First, we observe that removing distillation leads to a huge drop of performance.
This highlights the importance of our approach that consists in adding this memory distillation loss from a LBHT teacher model.
Then the performance increases until a plateau centered around $3 \times 10^{-5}$ while the performance starts decreasing again for too high values.

\vspace*{2mm}
\resizebox{\linewidth}{!}{
\begin{tabular}{l|l}
\hline
 Batch size          & 32 \\
 Optimizer           & AdamW ($\beta_1=0.9, \beta_2=0.99, \epsilon=1e{-}15$)\\
 Base Learning rate  & 1.5e-4 \\
 Minimum LR          & 1e-8 \\
 Scheduler           & Cosine annealing \\
 Warm-up epochs      & 40 \\
 Gradient clipping   & 1.0 \\
 Weight decay        & 0.05 \\
 Epochs              & 400 \\
 Weight for RPE loss & 1.0 \\
 Weight for Distillation loss &  $3 \times 10^{-5}$ \\
\hline
\end{tabular}
}

\begin{table}[t] 
    \centering
    \caption{\label{tab:wloss}\textbf{Impact of the weight $\lambda_{dist}$ of the distillation loss} on \rpeval. This ablation is done with training for 200 epochs. 
    }
\resizebox{0.9\linewidth}{!}{
\begin{tabular}{l|C{T200}C{T200}C{T200}|C{T800}C{T800}C{T800}}
            \specialrule{1pt}{0pt}{0pt}
            \rowcolor{TableGray2}
             \cellcolor{TableGray2} &
             \multicolumn{3}{c|}{\textbf{Seq len 200}} &
             \multicolumn{3}{c}{\textbf{Seq len 800}}
             \\
             \rowcolor{TableGray2}\textbf{$\lambda_{dist}$} 
             \cellcolor{TableGray2} \textbf{}&
             \metrics &
             \metrics
             \\[1pt]
             \specialrule{0.5pt}{0pt}{0pt}
            \cellcolor{TableGray2} $0.0$&
            5 & 23 & 47 & 3 & 15 & 37
            \\
             \cellcolor{TableGray2} $1 \times 10^{-5}$ &
            32 & 45 & 68 & 20 & 28 & 50 
             \\
             \cellcolor{TableGray2} $\bf{3 \times 10^{-5}}$ &
            \bf{38} & \bf{57} & 76 & \bf{25} & \bf{40} & \bf{63}  
             \\
             \cellcolor{TableGray2} $1 \times 10^{-4}$ &
            34 & \bf{57} & \bf{77} & 21 & 37 & 60
             \\
             \cellcolor{TableGray2} $3 \times 10^{-4}$ &
           36 & 56 & \bf{77} & 22 & 35 & 59
           \\
           \cellcolor{TableGray2} $1 \times 10^{-3}$ &
           27 & 49 & 74 & 16 & 30 & 54
           \\
            \specialrule{1pt}{0pt}{0pt}
\end{tabular}
}
\end{table}

\section{Mem-RPE distillation design choices}
\label{app:memrpe:distillation}

In Table~\ref{tab:ablationsdistilling} we report a sensitivity analysis of the distillation procedure into the student. We find that performing multi-step distillation improves performance significantly (\cf Equation~\ref{eq:distloss}), 
with performance saturating starting from $N{=}4$ segments. We also explore different backpropagation lengths and find that using a  length equal to the segment length performs well. This also allows for an efficient implementation, where at the end of each segment, memory is saved and detached and losses are applied for this segment. Finally, we study what happens if we also add a Masked Image Modeling loss like~\citep{kinaema2025} and find it detrimental to performance. We hypothesize that such an auxiliary loss helps training stability but does not improve the task itself. In our case, training stability is achieved thanks to memory distillation.

\begin{table}[t] \centering
    \caption{\label{tab:ablationsdistilling}\textbf{Design choices for distilling}: we explore different numbers $N$ of distillation segments (top) and limits for backpropagation (right, top rows, $N{=}4)$, as well as adding a Masked Image Modeling loss (bottom, bottom row, $N{=}4$). This ablation is performed under 200-epoch training schemes.}
\resizebox{0.8\linewidth}{!}{
\begin{tabular}{c|C{T200}C{T200}C{T200}|C{T800}C{T800}C{T800}}
            \specialrule{1pt}{0pt}{0pt}
            \rowcolor{TableGray2}
             \cellcolor{TableGray2}\textbf{Num.} &
             \multicolumn{3}{c|}{\textbf{Seq len 200}} &
             \multicolumn{3}{c}{\textbf{Seq len 800}}
             \\
             \rowcolor{TableGray2}
             \cellcolor{TableGray2} \textbf{segments}&
             \metrics &
             \metrics
             \\[1pt]
             \specialrule{0.5pt}{0pt}{0pt}
             \cellcolor{TableGray2} 1 & 
             14 & 33 & 61 & 8 & 19 & 45 \\
             \cellcolor{TableGray2} 2 & 
             16 & 41 & 66 & 10 & 25 & 50 \\
             \cellcolor{TableGray2} 3 & 
             23 & 49 & 72 & 15 & 34 & 58 \\
             \cellcolor{TableGray2} \bf{4} & 
             \bf{38} & \bf{57} & 76 & \bf{25} & \bf{40} & \bf{63} \\
             \cellcolor{TableGray2} 5 & 
             37 & \bf{57} & \bf{77} & \bf{25} & \bf{40} & \bf{63} \\
            \specialrule{1pt}{0pt}{0pt}
\end{tabular}
}
\\[0.1cm]
\resizebox{0.8\linewidth}{!}{
\begin{tabular}{cc|C{T200}C{T200}C{T200}|C{T800}C{T800}C{T800}}
            \specialrule{1pt}{0pt}{0pt}
            \rowcolor{TableGray2}
             \cellcolor{TableGray2} \textbf{Backprop}&
             \cellcolor{TableGray2} \textbf{+ MIM} &
             \multicolumn{3}{c|}{\textbf{Seq len 200}} &
             \multicolumn{3}{c}{\textbf{Seq len 800}}
              \\
             \rowcolor{TableGray2}
             \cellcolor{TableGray2} \textbf{length} &
             \cellcolor{TableGray2} \textbf{loss} &
             \metrics &
             \metrics
             \\[1pt]
             \specialrule{0.5pt}{0pt}{0pt}
             \cellcolor{TableGray2} 32& \cellcolor{TableGray2} no & 
             25 & 48 & 72 & 15 & 31 & 57 \\
             \cellcolor{TableGray2} \textbf{t/4} & \cellcolor{TableGray2} \textbf{no} & 
            \bf{38} & \bf{57} & \bf{76} & \bf{25} & \bf{40} & \bf{63} \\
             \cellcolor{TableGray2} 100 & \cellcolor{TableGray2} no & 
             20 & 46 & 71 & 12 & 28 & 53 \\
            \cellcolor{TableGray2} t/4 & \cellcolor{TableGray2} yes & 
            7 & 27 & 53 & 4 & 17 & 39 \\
            \specialrule{1pt}{0pt}{0pt}
            \multicolumn{7}{c}{~}
            \\
    \end{tabular}
}
\end{table}

\section{Mem-RPE wall-clock runtime measurements}

\begin{figure*}
    \centering
    \includegraphics[width=0.8\linewidth]{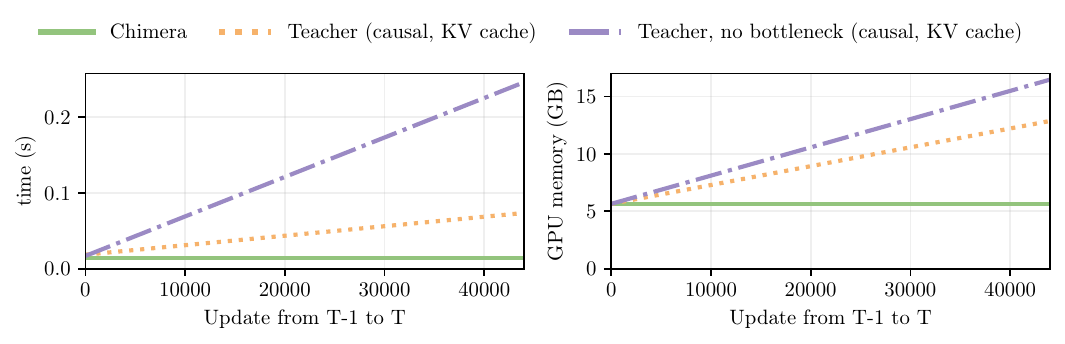}
    \caption{\textbf{Inference timings and memory requirements} for our \textit{\ours} model and a causal history transformer (attending over the full context) \textbf{\underline{with KV-cache}}.
    }
    \label{fig:timecausal}
\end{figure*}

\begin{figure*}
    \centering
    \includegraphics[width=0.8\linewidth]{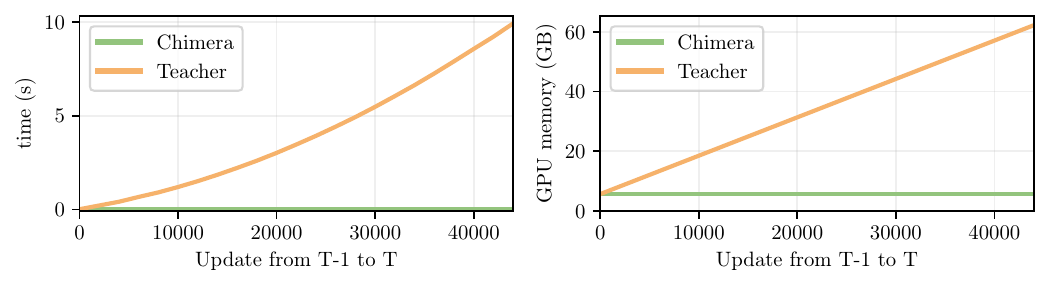}
    \vspace{-0.2cm}
    \caption{\textbf{Inference timings and memory requirements} for our \textit{\ours} model and a LBHT Teacher. 
    }
    \label{fig:time}
\end{figure*}

In order to showcase the benefit of recurrent transformers with $\mathcal{O}(1)$ update cost, compared to history transformers, we plot in~\Cref{fig:timecausal} and in~\Cref{fig:time} the runtime and GPU memory required to process sequences of increasing lengths. Timings are measured on a NVIDIA A100 GPU.

In ~\Cref{fig:timecausal} we compare Chimera to causal history transformers with KV-Cache (w/ and w/o bottlenecking layers), which have $O(T)$ complexity. In~\Cref{fig:time} we compare Chimera to the LBHT Teacher that exhibits quadratic complexity with $T$. In contrast, \textit{\ours} has a constant time of around 7ms between two timesteps.

Considering memory, \textit{\ours} has a constant usage, while History Transformers need to store the full history of frames, thus the linear complexity. 

\section{Mem-RPE qualitative examples}

\begin{figure*}
    \centering
    \begin{tikzpicture}
    \draw (-1, 0) node[anchor=west,inner sep=0] {
        \includegraphics[width=0.9\linewidth,trim={60pt 40pt 40pt 59pt},clip]{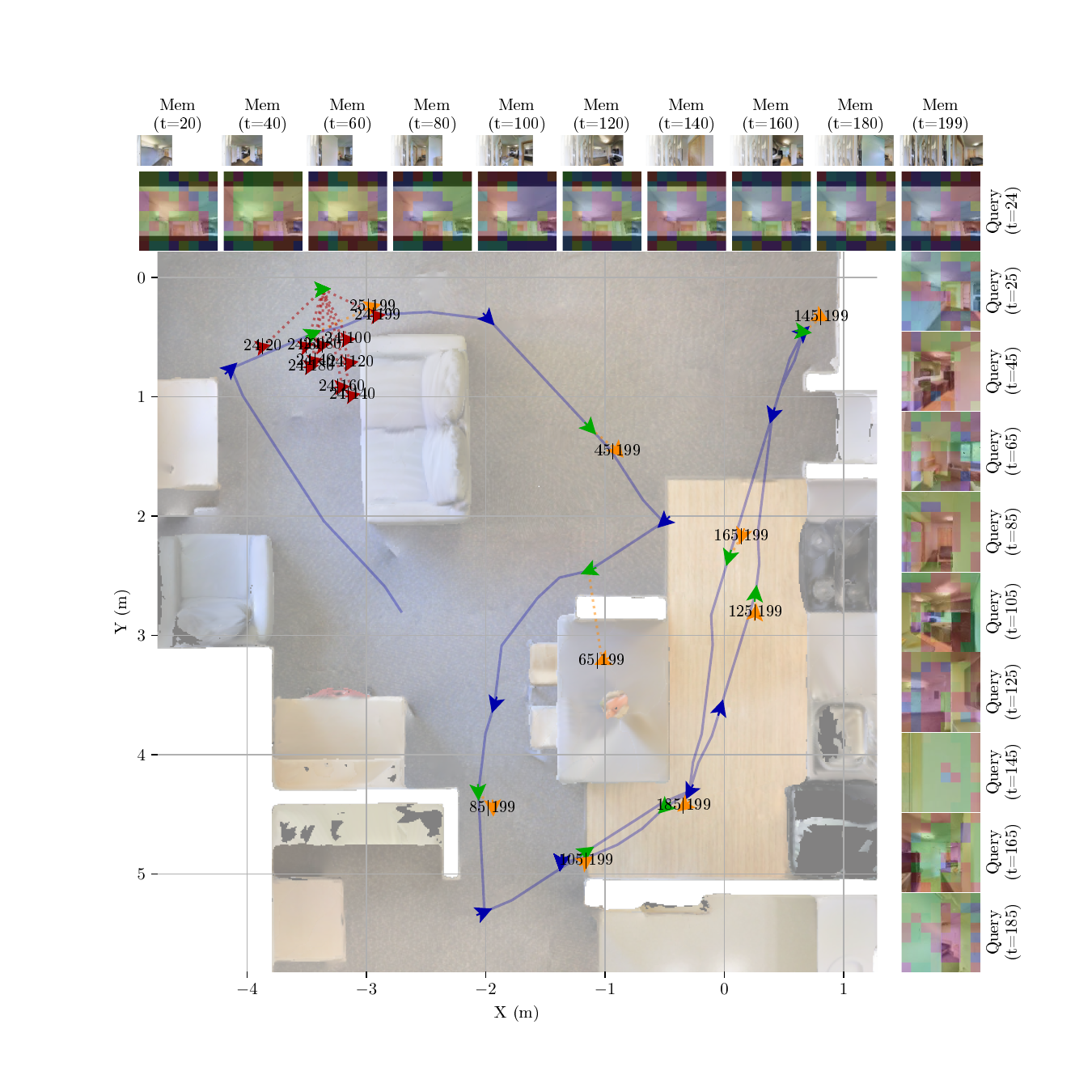}
    };
    \draw (-1.4, 7) node[anchor=west,inner sep=0] {\large \textcolor[HTML]{A91C11}{\ding{192}$\rightarrow$}};
    \draw (13.2, 8.2) node[anchor=west,inner sep=0] {\large \textcolor[HTML]{FE8D2B}{$\downarrow$\ding{193}}};
    \node[fill=white, 
              draw=black, 
              line width=0.5pt, 
              inner sep=5pt, 
              align=left,
              anchor=center,
              font=\scriptsize] at (1.0, -4.1) {
            \textbf{Legend:} \\
            \textcolor[HTML]{1700A6}{$\blacktriangleright$ trajectory followed} \\
            \textcolor[HTML]{00A822}{$\blacktriangleright$ GT pose (queried)} \\
            \textcolor[HTML]{A91C11}{$\blacktriangleright$ predictions from \ding{192}} \\
            \textcolor[HTML]{A91C11}{~~(Fixed $\mathbf{q}_t$, evolving $\mathbf{M}_t$)} \\
            \textcolor[HTML]{FE8D2B}{$\blacktriangleright$ predictions from \ding{193}} \\
            \textcolor[HTML]{FE8D2B}{~~(Fixed $\mathbf{M}_t$, evolving $\mathbf{q}_t$)}
        };
    \end{tikzpicture}
    \vspace{-0.4cm}
    \caption{Attention distributions over memory tokens for \textcolor[HTML]{A91C11}{\ding{192}$\longrightarrow$} a fixed query against an evolving memory, and \textcolor[HTML]{FE8D2B}{$\downarrow$\ding{193}} evolving queries against fixed memory extracted at the end.
    \label{fig:memrpe:attn}
    }
\end{figure*}

\begin{figure*}
    \centering
    \begin{tikzpicture}
    \draw (-1, 0) node[anchor=west,inner sep=0] {
        \includegraphics[width=\linewidth]{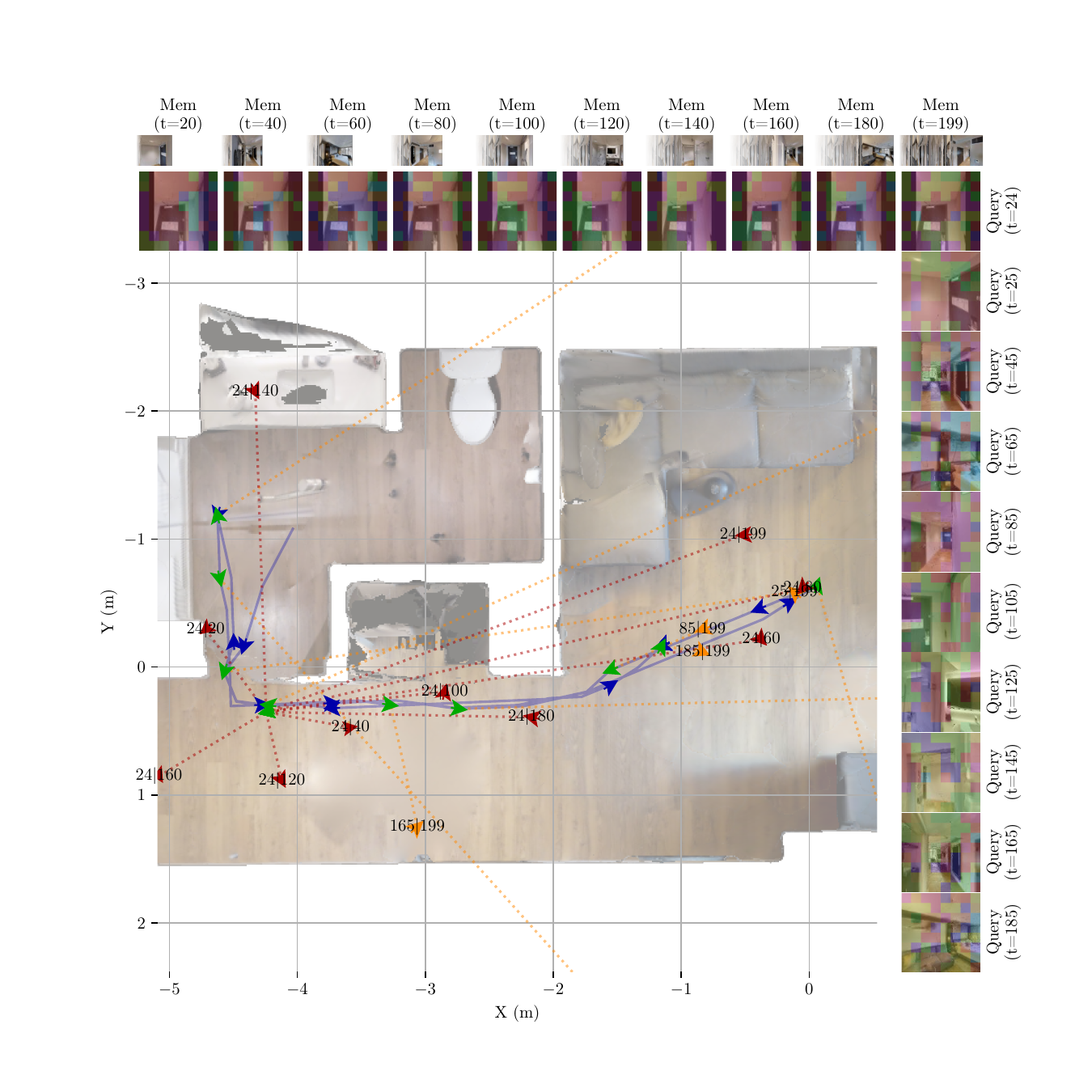}
    };
    \draw (0, 6.3) node[anchor=west,inner sep=0] {\large \textcolor[HTML]{A91C11}{\ding{192}$\rightarrow$}};
    \draw (14, 7.7) node[anchor=west,inner sep=0] {\large \textcolor[HTML]{FE8D2B}{$\downarrow$\ding{193}}};
    \node[fill=white, 
              draw=black, 
              line width=0.5pt,
              inner sep=5pt, 
              align=left,
              anchor=center,
              font=\scriptsize] at (1.0, -4.1) {
            \textbf{Legend:} \\
            \textcolor[HTML]{1700A6}{$\blacktriangleright$ trajectory followed} \\
            \textcolor[HTML]{00A822}{$\blacktriangleright$ GT pose (queried)} \\
            \textcolor[HTML]{A91C11}{$\blacktriangleright$ predictions from \ding{192}} \\
            \textcolor[HTML]{A91C11}{~~(Fixed $\mathbf{q}_t$, evolving $\mathbf{M}_t$)} \\
            \textcolor[HTML]{FE8D2B}{$\blacktriangleright$ predictions from \ding{193}} \\
            \textcolor[HTML]{FE8D2B}{~~(Fixed $\mathbf{M}_t$, evolving $\mathbf{q}_t$)}
        };
    \end{tikzpicture}
    \vspace{-0.4cm}
    \caption{Attention distributions over memory tokens for \textcolor[HTML]{A91C11}{\ding{192}$\longrightarrow$} a fixed query against an evolving memory, and \textcolor[HTML]{FE8D2B}{$\downarrow$\ding{193}} evolving queries against fixed memory extracted at the end.
    \label{fig:memrpe:attn:sm1}
    }
\end{figure*}

\begin{figure*}
    \centering
    \begin{tikzpicture}
    \draw (-1, 0) node[anchor=west,inner sep=0] {
        \includegraphics[width=\linewidth]{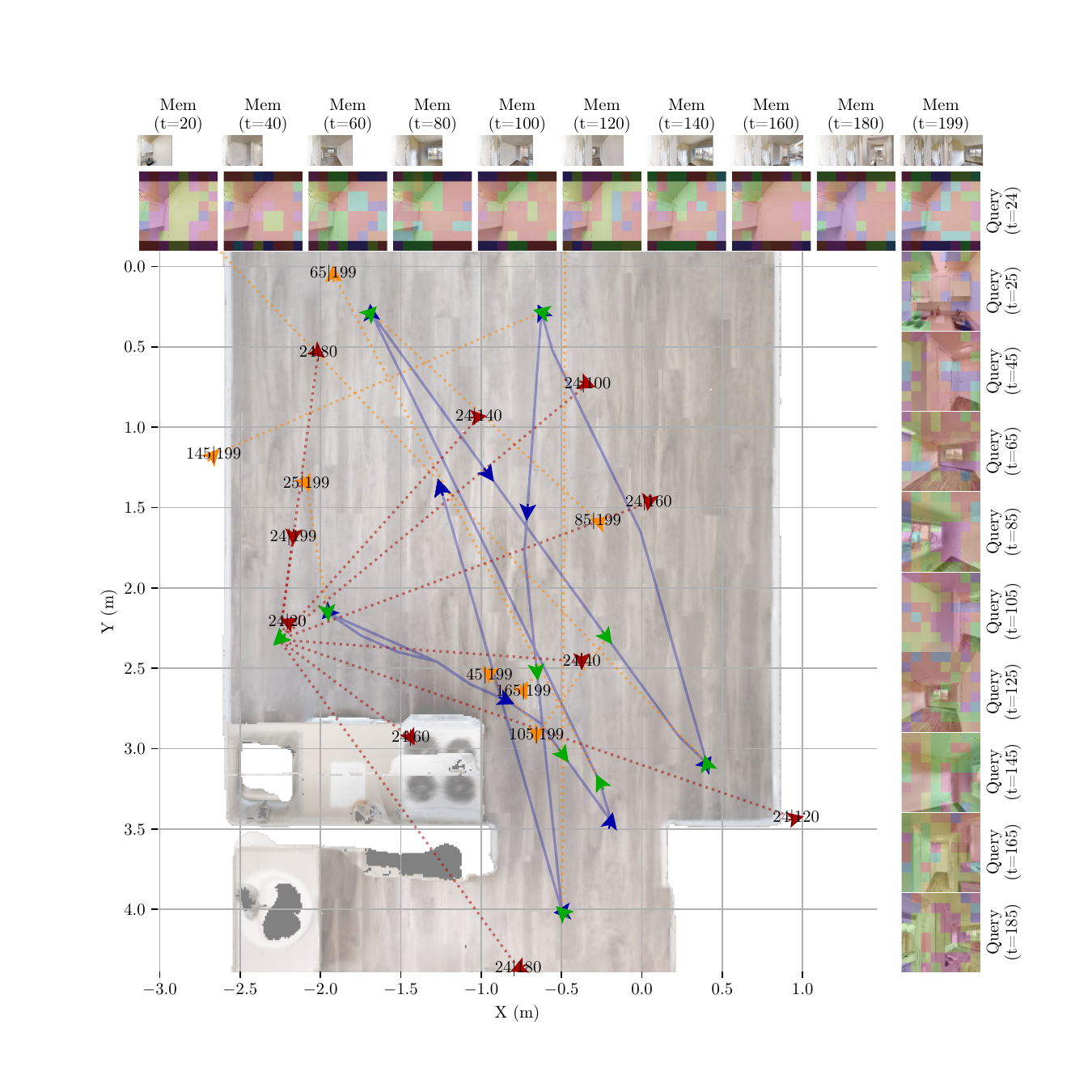}
    };
    \draw (-0.7, 3.8) node[anchor=west,inner sep=0] {\large \textcolor[HTML]{A91C11}{\ding{192}$\rightarrow$}};
    \draw (10.2, 4.5) node[anchor=west,inner sep=0] {\large \textcolor[HTML]{FE8D2B}{$\downarrow$\ding{193}}};
    \node[fill=white, 
              draw=black, 
              line width=0.5pt, 
              inner sep=5pt, 
              align=left,
              anchor=center,
              font=\scriptsize] at (1.0, -4.1) {
            \textbf{Legend:} \\
            \textcolor[HTML]{1700A6}{$\blacktriangleright$ trajectory followed} \\
            \textcolor[HTML]{00A822}{$\blacktriangleright$ GT pose (queried)} \\
            \textcolor[HTML]{A91C11}{$\blacktriangleright$ predictions from \ding{192}} \\
            \textcolor[HTML]{A91C11}{~~(Fixed $\mathbf{q}_t$, evolving $\mathbf{M}_t$)} \\
            \textcolor[HTML]{FE8D2B}{$\blacktriangleright$ predictions from \ding{193}} \\
            \textcolor[HTML]{FE8D2B}{~~(Fixed $\mathbf{M}_t$, evolving $\mathbf{q}_t$)}
        };
    \end{tikzpicture}
    \vspace{-0.4cm}
    \caption{Attention distributions over memory tokens for \textcolor[HTML]{A91C11}{\ding{192}$\longrightarrow$} a fixed query against an evolving memory, and \textcolor[HTML]{FE8D2B}{$\downarrow$\ding{193}} evolving queries against fixed memory extracted at the end.
    \label{fig:memrpe:attn:sm2}
    }    
\end{figure*}

\begin{figure*}
    \centering
    \begin{tikzpicture}
    \draw (-1, 0) node[anchor=west,inner sep=0] {
        \includegraphics[width=\linewidth]{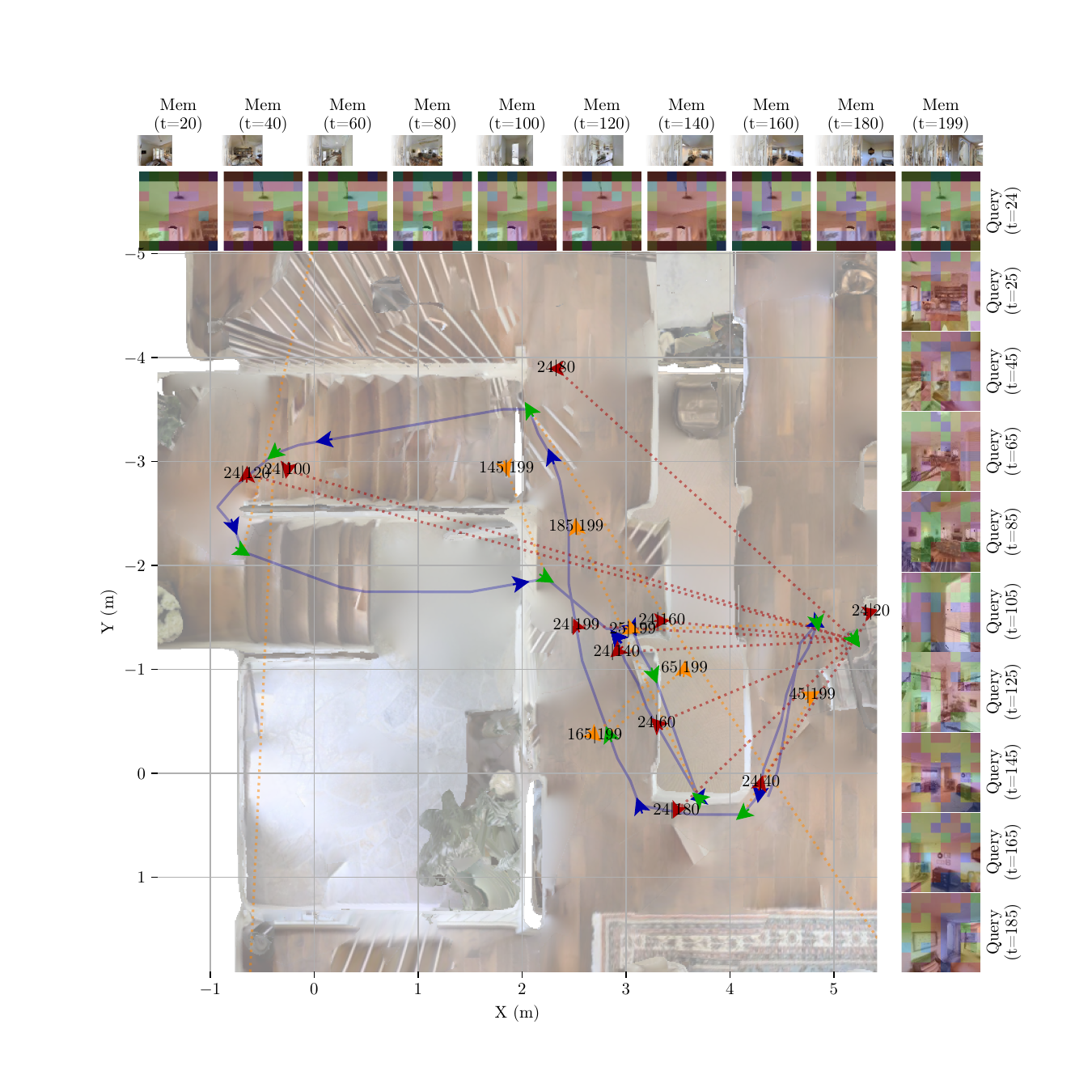}
    };
    \draw (-0.7, 3.8) node[anchor=west,inner sep=0] {\large \textcolor[HTML]{A91C11}{\ding{192}$\rightarrow$}};
    \draw (10.2, 4.5) node[anchor=west,inner sep=0] {\large \textcolor[HTML]{FE8D2B}{$\downarrow$\ding{193}}};
    \node[fill=white, 
              draw=black, 
              line width=0.5pt,
              inner sep=5pt, 
              align=left,
              anchor=center,
              font=\scriptsize] at (1.0, -4.1) {
            \textbf{Legend:} \\
            \textcolor[HTML]{1700A6}{$\blacktriangleright$ trajectory followed} \\
            \textcolor[HTML]{00A822}{$\blacktriangleright$ GT pose (queried)} \\
            \textcolor[HTML]{A91C11}{$\blacktriangleright$ predictions from \ding{192}} \\
            \textcolor[HTML]{A91C11}{~~(Fixed $\mathbf{q}_t$, evolving $\mathbf{M}_t$)} \\
            \textcolor[HTML]{FE8D2B}{$\blacktriangleright$ predictions from \ding{193}} \\
            \textcolor[HTML]{FE8D2B}{~~(Fixed $\mathbf{M}_t$, evolving $\mathbf{q}_t$)}
        };
    \end{tikzpicture}
    \vspace{-0.4cm}
    \caption{Attention distributions over memory tokens for \textcolor[HTML]{A91C11}{\ding{192}$\longrightarrow$} a fixed query against an evolving memory, and \textcolor[HTML]{FE8D2B}{$\downarrow$\ding{193}} evolving queries against fixed memory extracted at the end.
    \label{fig:memrpe:attn:sm3}
    }        
\end{figure*}

\Cref{fig:memrpe:attn,fig:memrpe:attn:sm1,fig:memrpe:attn:sm2,fig:memrpe:attn:sm3} provide Mem-RPE qualitative examples. 
They show the evolution of attention patterns of \textit{\ours}{}  for one trajectory per Figure, for \textcolor[HTML]{A91C11}{\ding{192}$\rightarrow$} a fixed query against an evolving memory, and \textcolor[HTML]{FE8D2B}{$\downarrow$\ding{193}} evolving queries against fixed memory extracted at the end. There does not seem to be a strong correlation between memory tokens and scene semantics, \eg walls, floor, windows can trigger different tokens in different trajectories. We observe some stability and coherence along a specific trajectory, as dominant memory tokens will follow the motion of the camera, and a fixed query will also mostly conserve its attention patterns against an evolving memory. It is also worth noting that some tokens seem to receive more attention overall, as if the model had ``default registers'' regardless of the input observations. 

\section{Details on Streaming Visual Question Answering}
\label{app:vqa}

This section provides the architecture, data, and training details for the video-language application (\Cref{sec:vqa}). 
These experiments complement our main Mem-RPE study by testing the same principle in a different prediction setting: first learn to compress an accessible observation history into a bottleneck, then use this representation to supervise the hidden state of a recurrent model.

\subsection{Visual representations and model adaptation}
\label{sec:vqa:architecture}

\begin{figure*}
    \centering
    Teacher architecture illustration \\
     \includegraphics[width=0.8\linewidth]{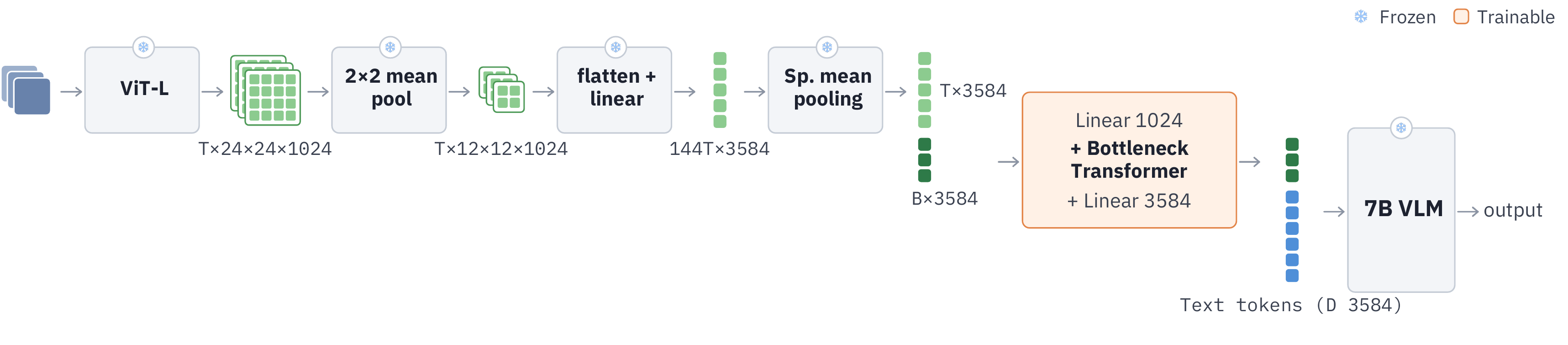} \\
    Student architecture illustration \\
    \includegraphics[width=0.8\linewidth]{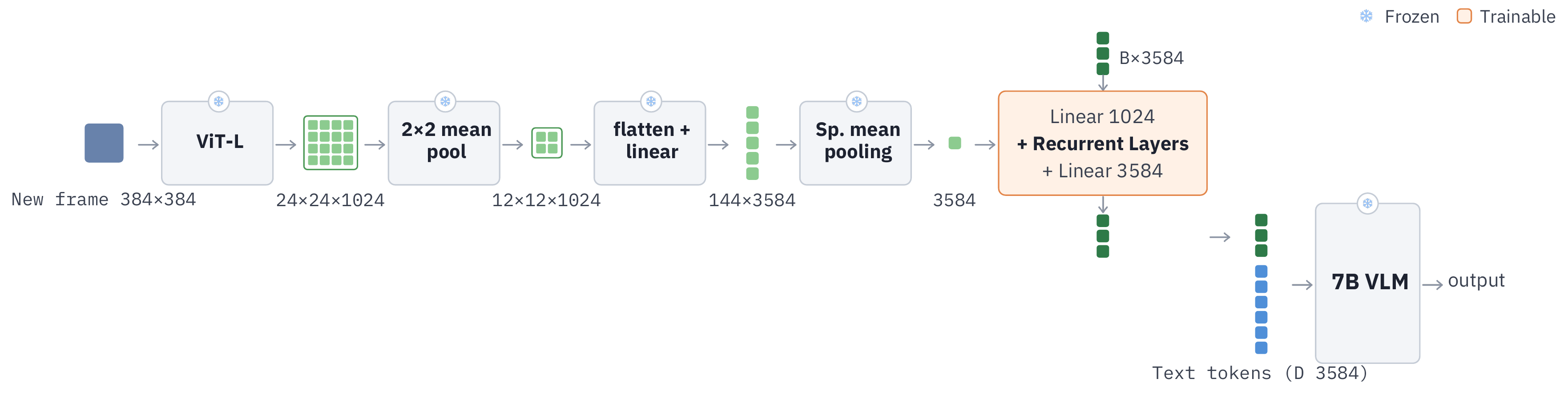}
    \vspace{-0.3cm}
     \caption{Illustration of the teacher (top) and student (bottom) architectures for Streaming Visual Question Answering.}
     \label{fig:vqa}
\end{figure*}

We start from LongVA-7B~\citep{longva}, whose language model has been trained for long contexts and has been shown to transfer well for processing long video inputs. Its ViT-Large visual encoder, $2\times2$ spatial pooling, and projection to 3584 dimensions (the one from the language model) produce 144 visual tokens per frame from video inputs resized to $384{\times}384$px. 
In our experiments, we average these tokens spatially to obtain one global representation per frame. This additional pooling reduces the input to the temporal model to one token per observed frame. The visual encoder and language model remain frozen during teacher and student training.

The teacher uses six non-causal transformer layers of width 1024. It processes the frame tokens together with $B{=}32$ learnable bottleneck tokens, whose contextualized outputs provide the visual representation for the language model. The recurrent student processes the frame representations one by one and replaces joint history processing with a fixed-size memory update. In both cases, temporal compression is independent of the query \ie of the question. The resulting teacher or student representation is mapped to the language model's input space and combined with the textual question for answer prediction. Figure~\ref{fig:vqa} summarizes these two paths.

\subsection{Data and temporal sampling}

\myparagraph{Training data.} We train on LLaVA-Video~\citep{llavavideo}, a collection of 178,510 videos with video-language annotations. The videos' duration ranges from 5 to 180 seconds. Frame features are extracted at 6 fps, and training uses a random stride $s\in\{1,\ldots,12\}$ over the cached features. The effective frame rate thus ranges from 0.5 to 6 fps. At the highest sampling rate, a three-minute video contains up to 1080 frame tokens. A subset of LLaVA-Video is kept for validation using multiple-choice questions.

\myparagraph{Evaluation data.}
Streaming evaluation uses OVO-Bench~\citep{ovobench} and the `Real' task of StreamingBench~\citep{streambench} (the `real' subset mainly focuses on visual content; the other tasks involve sequential question answering where later questions refer to entities or events encountered earlier, with the `omni' subset mostly based on audio). 
Predictions use the video observations available at query time.  
We also report offline results on the held-out LLaVA-Video multiple-choice subset and on Video-MME~\citep{videomme}. In the offline setting, a complete video is available before answering, whereas the streaming setting requires answers from successive prefixes.

\subsection{Training the teacher and student}
\label{sec:vqa:training}

The teacher is first trained with the vision-language answer-prediction loss. The frozen language model provides supervision for learning which visual information to retain in the bottleneck. The teacher's visual history encoder is non-causal over the available prefix, but cannot access observations beyond the endpoint being supervised.

Student training follows the multi-segment distillation strategy. For a video represented by $T$ frame tokens, we use $N{=}4$ endpoints, approximately $e_n=nT/4$ for $n\in\{1,2,3,4\}$. The frozen teacher is run independently on each prefix $\{\obs_1,\ldots,\obs_{e_n}\}$ to obtain its bottleneck representation. One sequential student rollout supplies the corresponding recurrent memory states. We combine the VQA loss with $L_1$ matching of the aligned teacher and student representations, as in Equation~\ref{eq:distloss} and~\ref{eq:studentloss}. 
Compared to the Mem-RPE setting, the task loss is only applied at the final step.
The baseline without distillation uses the same recurrent architecture and VQA supervision, with $\lambda_{dist}=0$.

\subsection{Teacher ablations}

Table~\ref{tab:vqateacher} reports performance for various LBHT configurations, varying the number of bottleneck tokens $B$, and the transformer depth $d$. 
We first report a no-vision version, where the frozen vision-language model input is limited to text tokens.
We also consider a transformer without bottleneck, for which the language model receives the processed tokens corresponding to all input frames.

On the held-out LLaVA-Video subset, the non-bottleneck transformer achieves $73.4\%$, compared with $72.4\%$ for the default $B{=}32$, $d{=}6$ teacher. Bottlenecks with 8 and 16 tokens achieve $71.9\%$ and $72.2\%$, respectively.
Further increasing the number of bottleneck tokens or the transformer depth does not bring noticeable gains.
We use the 6-layer, 32-token non-causal model as the distillation teacher.

Although the bottleneck limits the number of visual tokens passed to the language model, its history transformer still processes a sequence whose length grows with the video.
In streaming use-cases, non-causal history representations must be recomputed for successive prefixes; the frozen per-frame features can be cached, but the temporal transformer cannot reuse its previous contextualized representations in the same way as a causal KV cache. In contrast, the recurrent student carries its fixed-size visual memory forward. This provides bounded visual-history storage and constant-cost memory updates at a fixed model size; it does not establish an offline wall-clock speed advantage, since a history transformer can process a complete video in parallel.

\begin{table*}
    \centering
    \caption{\textbf{Bottleneck teacher ablations on video question answering} on the number of bottleneck tokens (B) and the depth (d) of the transformer. Results are reported on both non-streaming (LLaVA-Video val \& Video-MME) as well as streaming benchmarks (OVO-Bench, StreamingBench Real task).}
    \setlength{\tabcolsep}{3pt}
    \begin{tabular}{lcccc}
    \specialrule{1pt}{0pt}{0pt}
    \rowcolor{TableGray2}
    \cellcolor{TableGray1}
    Model & Llava-Video & Video-MME & OVO-Bench & StreamingBench \\
    \rowcolor{TableGray2}
    \cellcolor{TableGray1} & (val) & & & (real) \\
    \specialrule{0.5pt}{0pt}{0pt}
    \cellcolor{TableGray1} no vision          & 57.1 & 37.1 & 33.5 & 39.3 \\ 
    \cellcolor{TableGray1} non-bottleneck transformer (d=6) & 73.4 & 38.8 & 35.5 & 50.3 \\
    \cellcolor{TableGray1} LBHT (B=8,d=6) & 71.9 & 41.1 & 34.9 & 50.3 \\ 
    \cellcolor{TableGray1} LBHT (B=16,d=6) & 72.2 & 41.6 & 35.6 & 48.6 \\ 
    \cellcolor{TableGray1} \bf{LBHT (B=32,d=6)} & 72.4 & 40.6 & 37.0 & 49.4 \\
    \cellcolor{TableGray1} LBHT (B=64,d=6) & 71.7 & 41.8 & 34.5 & 50.0 \\ 
    \cellcolor{TableGray1} LBHT (B=32,d=9) & 71.2 & 40.7 & 36.1 & 49.5 \\
    \specialrule{1pt}{0pt}{0pt}
    \end{tabular}
    \label{tab:vqateacher}
\end{table*}

\end{document}